\documentclass[12pt]{article}
\usepackage{amsmath}
\usepackage{graphicx,psfrag,epsf}
\usepackage{enumerate}
\usepackage{natbib}
\usepackage{url} % not crucial - just used below for the URL
\usepackage{algorithm}
\usepackage{algpseudocode}
\usepackage{tabularray}
\usepackage{multirow}
\usepackage{comment}
\usepackage{enumitem}
\usepackage{authblk}
\usepackage{caption}
\usepackage{tikz}
\usepackage{threeparttable}

\usepackage{hyperref}       % hyperlinks
\usepackage{url}            % simple URL typesetting
\usepackage{booktabs}       % professional-quality tables
\usepackage{amsfonts}       % blackboard math symbols
\usepackage{mathrsfs}
\usepackage{nicefrac}       % compact symbols for 1/2, etc.
\usepackage{microtype}      % microtypography
\usepackage{xcolor}         % colors
\usepackage{graphicx}
\usepackage{epstopdf}
\usepackage{float}
\usepackage{subfigure}
\usepackage{amssymb}
\usepackage{adjustbox}
\usepackage{bbding}
\usepackage{amsmath}
\usepackage{setspace}
\usepackage{color}
\usepackage{pifont}

\newtheorem{theorem}{Theorem}[section]% meant for sectionwise numbers
\newtheorem{corollary}{Corollary}[section]% meant for sectionwise numbers

\newtheorem{remark}{Remark}%
\newtheorem{definition}{Definition}

\newtheorem{assumption}{Assumption}
\newtheorem{lemma}{Lemma}[section]

\newcommand{\blind}{0}

\usepackage{hyperref}
\hypersetup{
    colorlinks=true,       % true = colored text; false = boxed links
    linkcolor=blue,        % color of internal section/figure links
    filecolor=magenta,     % color of file links
    urlcolor=cyan,         % color of external website links
    citecolor=blue,       % color of citation links
}

\begin{document}

\def\spacingset#1{\renewcommand{\baselinestretch}%
{#1}\small\normalsize} \spacingset{1}

\if0\blind
{
  \title{\bf  Semi-Supervised Conditional Diffusion via Label Augmentation}
\author{Jin $\rm{Su}^{a}$,  Yuan $\rm{Gao}^{b}$,
	Yong $\rm{Zhou}^{c}$, and
	Jian $\rm{Huang}^{a,d}$\thanks{Corresponding author.}\\
	\vspace{-0.2cm}
	{\small $^{a}$Department of Applied Mathematics, The Hong Kong Polytechnic University, Hong Kong, China}\\
\vspace{-0.2cm}
{\small $^{b}${%LPMC \& KLMDASR,
School of Statistics and Data Science}, Nankai University, Tianjin, China} \\
\vspace{-0.2cm}
{\small $^{c}$School of Statistics, %Academy of Statistics and Interdisciplinary Sciences,
East China Normal University, Shanghai, China}\\
%Key Laboratory of Advanced Theory and Application in Statistics and Data Science MOE,
\vspace{-0.2cm}
{\small $^{d}$Department of Data Science and AI, The Hong Kong Polytechnic University, Hong Kong, China}
}
\date{\footnotesize \today}
  \maketitle} \fi

\if1\blind
{
  \bigskip
  \bigskip
  \bigskip
  \begin{center}
    {\bf Semi-Supervised Conditional Diffusion via Label Augmentation}
\end{center}
  \medskip
} \fi

\bigskip
	\abstract{Conditional diffusion models have become a powerful and flexible framework for learning complex conditional distributions from labeled data. In practice, however, acquiring high-quality labels is costly and time-consuming, leaving large volumes of unlabeled data unused. To address this, we introduce label‑augmented conditional diffusion (LACD), a simple and effective approach that incorporates unlabeled examples by assigning them a designated trivial label and performing joint denoising score matching over the augmented dataset. We provide sufficient conditions guaranteeing population‑level identifiability of the target conditional distribution under this scheme. Moreover, we establish rigorous statistical guarantees: when sufficiently many unlabeled samples are available, the sampling distribution produced by LACD converges strictly faster than the purely supervised estimator in total variation distance, and at least as fast in Wasserstein‑1 distance. Extensive experiments on synthetic, image, and tabular benchmarks corroborate our theory and show substantial gains in sample efficiency and generative performance compared with the purely supervised estimator.}

\bigskip
\noindent
{\it Keywords:} {Semi-supervised learning, diffusion models, conditional distribution, generative modeling, shared embedding}
\vfill

\newpage
\spacingset{1.5} % DON'T change the spacing!
	\section{Introduction}
	\label{sec:Introduction}
	Estimating the conditional distribution is a central problem in statistics and machine learning.
	Beyond finite-dimensional functionals (e.g. summary statistics or regression parameters) and predictive objects (e.g. conditional mean or quantile), modeling the distribution can capture more complex data structures.
	This capability drives modern advancements in representation learning \citep{sohn2015learning,khemakhem20a,preechakul2022diffusion}, data augmentation \citep{shin2018medical, tian2025conditional}, and privacy-preserving synthetic data generation \citep{beaulieu2019privacy, daum2024differentially}.
	Furthermore, it provides a principled  approach for resolving statistical problems, including missing data imputation \citep{yoon2018gain, mattei2019miwae, ipsen2021notmiwae}, counterfactual reasoning \citep{wu2024counterfactual,wu2025po,chen2025enhancingcausaleffectestimation}, and nonparametric regression \citep{song2025wasserstein}.
	
Let $(X, Y)$ be a pair of random variables. The existing methods, including kernel smoothing \citep{nadaraya1964estimating,watson1964smooth,wasserman2006all} and other parametric and nonparametric approaches \citep{mclachlan2019finite, chacon2018multivariate}, seek to estimate the functional form of the conditional density of $Y$ given $X.$
%P_{X|Y}$ \citep{zhou2023deep}.
	However,% in addition to the curse of dimensionality,
in multivariate and high-dimensional settings,
these methods exhibit high estimation variance when labeled samples are scarce and are inadequate for characterizing the complex, multimodal distributions encountered in modern applications.
	While generative approaches, such as conditional diffusion models \citep{ho2021classifierfree,nichol2022glide,rombach2022high,bansal2023universal,zhang2023adding}, relax these structural restrictions and achieve minimax optimality under the supervised setting \citep{fu2024unveil, tang2024conditional}, their performance remains limited by the available paired data. Consequently, when paired data  $(X,Y)$ is scarce, these models suffer from overfitting \citep{giannone2022fewshot,carlini2023extracting,jeon2025understanding}.
	
	To mitigate the dependence on labeled data, semi-supervised learning (SSL) offers a natural solution. In practice, acquiring paired observations %$(X, Y)$
is often expensive or restricted by privacy concerns, whereas unlabeled data are abundant and readily available. This data imbalance motivates an SSL setting with
	a labeled dataset $\mathcal{D}_l = \{(X_i, Y_i)\}_{i=1}^{N_{\ell}}$ and a
	larger unlabeled dataset $\mathcal{D}_u = \{X_i\}_{i=N_{\ell}+1}^{N}$, where
	$N_u = N - N_{\ell} \gg N_{\ell}$.
	The theoretical properties of SSL have been extensively explored for discriminative tasks, particularly in finite-dimensional parameter estimation \citep{cai2018efficient, zhang2019mean, zhang2021mean, azriel2022semi, angelopoulos2023prediction, deng2024optimal, song2024general,kallus2025role} and nonparametric regression \citep{wasserman2007statistical,ding2025semisupervised}.
In those works, unlabeled observations are usually used for $Y$-related targets: $X$ is treated as covariates, $Y$ is the response or label, and the goal is typically the conditional distribution $P_{Y|X}$ or a $Y$-related functional of this conditional distribution.  A key insight is that unlabeled observations improve statistical efficiency when the marginal distribution provides structural information for target estimands.
	
Our problem has the same observed data structure but a different target.
We study semi-supervised conditional generation, where $X$ is the object to be generated and the target is $P_{X|Y}$.
		Thus, unlabeled observations of $X$ provide direct information about the marginal distribution of the generated object, while the labeled pairs identify how this distribution varies with $Y$.
	Several related lines of work address neighboring aspects of this problem but do not cover this setting. Semi-supervised or auxiliary-data methods for distributional estimation provide efficiency gains for marginal densities or conditional distribution functions \citep{cannings2022correlation,mengtaowen2024}.
	Related generative-learning work studies transfer learning from source models \citep{tian2025enhancingaccuracygenerativemodels} or empirical semi-supervised diffusion through pseudo-labeling \citep{NEURIPS2023_8735753c}.
	Despite these developments, existing work leaves open how unlabeled observations of $X$ can be used to improve estimation of $P_{X|Y}$ in conditional generation.

\iffalse
	To address this challenge, recent research has increasingly focused on leveraging auxiliary data to improve estimation efficiency.
	\citet{cannings2022correlation} propose a correlation-assisted kernel density estimator for the marginal distribution under a setting where certain features are missing completely at random.
	\citet{mengtaowen2024} develop a bias-amended framework for the conditional cumulative distribution. Their approach employs bias-correction and cross-fitting to achieve uniform estimation efficiency.
	While these approaches provide rigorous efficiency gains in addressing incomplete observations, they are subject to the curse of dimensionality and are limited in their ability to capture complex distributions.
	In generative modeling, \cite{tian2025enhancingaccuracygenerativemodels} establish convergence rates under a transfer learning setting, demonstrating that pre-trained source models can enhance target generation accuracy via shared latent embeddings.
	However, for conditional tasks, this transfer mechanism requires a fully labeled and well-aligned source dataset.
	While recent empirical work has explored SSL for diffusion models via pseudo-labeling \citep{NEURIPS2023_8735753c}, theoretical properties remain largely unexplored.
\fi	

	% In this work, we propose a label-augmentation method for learning conditional distribution through a diffusion model in the semi-supervised learning setting.
	In this work, we propose a label-augmentation method for learning conditional distributions  with diffusion models in a semi-supervised setting.
	We assume that the conditional distributions $P_{X|Y=y}$
	share a common baseline component across all conditions $y\in\mathcal{Y}$.
	This allows unlabeled data to inform estimation of the shared component, while labeled pairs identify the condition-specific components.
	To implement this,
	unlabeled observations are assigned a trivial label, augmenting the conditioning space to $\widetilde{\mathcal{Y}} = \mathcal{Y} \cup \{\varnothing\}$ and integrating the marginal distribution and the conditional distributions within a single probability space. Here $\varnothing$ corresponds to unlabeled $X.$
	We employ score-based diffusion models \citep{song2019generative, song2020score}, where the score function is estimated via denoising score matching on the augmented dataset, and samples from $P_{X|Y=y}$ are obtained by simulating the reverse process conditioned on $y \in \mathcal{Y}$.
	Finite-sample convergence rates in total variation (TV) and Wasserstein-1 ($W_1$) distances are established, showing strict improvement over the supervised counterpart in TV distance, and in $W_1$ distance when the covariate structure is not too simple. These theoretical results are corroborated by extensive empirical studies across synthetic and real datasets.
	
The remainder of this paper is organized as follows. Section \ref{sec:Methodology} presents the methodology, Section \ref{sec:convergence_analysis} establishes convergence analysis, Section \ref{sec:Numerical_Experiments} reports experimental results, and Section \ref{sec:Conclusion} concludes.
%The supplementary materials provide theoretical proofs and experimental details.

\section{Methodology}
	\label{sec:Methodology}
	
	% In this section, we detail the proposed Label-Augmented Conditional Diffusion (LACD) Model.
	In this section, we describe the proposed Label-Augmented Conditional Diffusion (LACD) model.
	
	\subsection{Problem Setup and Augmented Probability Space}
Let $X \in \mathcal{X}\subseteq \mathbb{R}^{d_x}$ and $Y\in \mathcal{Y}=\{0,1,\ldots,K-1\}$ be the covariate and label, respectively.
Consider $N$ samples where only a small fraction are labeled.
Let $D \in \{0,1\}$ be a missing indicator, where $D=1$ if the label is observed and $D=0$ otherwise.
We assume that $D$ is independent of $(X,Y)$ with observation probability $\pi_N = \mathbb{P}(D=1).$
Define an augmented variable $\widetilde{Y} \in \widetilde{\mathcal{Y}} := \mathcal{Y} \cup \{\textcolor{red}{\varnothing}\}$ by
\begin{align*}
	\widetilde{Y} :=
	\begin{cases}
		Y, & D=1,\\
		\varnothing, & D=0,
	\end{cases}
\end{align*}
where $\varnothing$ represents the unlabeled state.
Our dataset $\mathcal{D} = \{(X_i, \widetilde{Y}_i)\}_{i=1}^N$ consists of $N$ independent observations from the joint distribution of $(X, \widetilde{Y})$.
This naturally partitions into a labeled dataset $\mathcal{D}_l = \{(X_i, Y_i) : \widetilde{Y}_i = Y_i \in \mathcal{Y}\}$ with size $N_{\ell}$, and an unlabeled dataset $\mathcal{D}_u = \{(X_i, \varnothing) : \widetilde{Y}_i = \varnothing\}$ with size $N_u$, where $N_{\ell} + N_u = N$. Our goal is to learn the conditional distribution $P_{X|Y}$ from observed data.

\begin{remark}
	In the missing data literature, missing completely at random (MCAR) is defined as the indicator $D$ being independent of $(X, Y)$, i.e., $P(D=1 \mid X, Y) = P(D=1)$ \citep{little2019statistical}.
	While MCAR settings typically assume ``positive overlap'' with a fixed labeling probability $P(D=1) = c > 0$, SSL settings
	generalize this by introducing a vanishing label proportion.
	Specifically, SSL settings allows $N_{\ell}/N \to 0$ (equivalently $\pi_N=P(D=1)  \to 0$) as $N \to \infty$, provided that $N \pi_N \to \infty$~\citep{song2024general,kallus2025role}.
	This setup ensures labeled data availability in finite samples and asymptotic growth in absolute size despite vanishing relative proportion, with generality to revert to balanced cases via fixed $\pi_N = c > 0$.
\end{remark}

Under the SSL setting described above, it can be shown that the conditional distribution $P_{X|\widetilde{Y}=\tilde{y}}$  satisfies
\begin{align*}
	P_{X|\widetilde{Y}=\tilde{y}} = \begin{cases} P_{X|Y=y} &\text{if } \tilde{y} = y \in \mathcal{Y}, \\ P_X & \text{if }\tilde{y} = \varnothing, \end{cases}
\end{align*}
so that the conditional probability $ P_{X|\widetilde{Y}=\tilde{y}}$ in the augmented space
identifies $P_{X|Y}$ and $P_X$, respectively.
We establish this identification result through a measure-theoretic construction.

Consider the original probability space $(\Omega, \mathcal{F}, P)$ with random variables $X(\omega): \Omega \to \mathcal{X}$ and $Y(\omega): \Omega \to \mathcal{Y}$. The joint distribution of $(X,Y)$ is $P \circ (X,Y)^{-1} \in \mathcal{P}(\mathcal{X} \times \mathcal{Y})$, where $\mathcal{P}(\mathcal{X} \times \mathcal{Y})$ denotes the set of all probability measures defined on $\mathcal{X} \times \mathcal{Y}$.

We construct an augmented probability space through a product measure space
$$(\Omega^*, \mathcal{F}^*, P^*) = (\Omega \times \{0,1\}, \mathcal{F} \otimes 2^{\{0,1\}}, P \otimes \mu_N),$$
 where $\mu_N$  is a Bernoulli measure on $\{0,1\}$ with $\mu_N({1}) = \pi_N$. For $\omega^* = (\omega, d) \in \Omega^*$ with $d\in\{0,1\}$, let  $X^*(\omega^*) = X(\omega)$, $Y^*(\omega^*) = Y(\omega)$  and $D(\omega^*)  = d$. The joint distribution of $(X^*,Y^*)$ is $ P^* \circ (X^*,Y^*)^{-1}$. This augmentation preserves the joint distribution of $(X,Y)$, as formalized below.
\begin{lemma}
	\label{lem:aug_joint_distribution}
	$ P^* \circ (X^*,Y^*)^{-1} = P \circ (X,Y)^{-1} \in \mathcal{P}(\mathcal{X} \times \mathcal{Y})$.	
\end{lemma}
Define the augmented variable $\widetilde{Y}: \Omega^* \to
	\widetilde{\mathcal{Y}}$ as $\widetilde{Y}(\omega^*) =  Y^*(\omega^*)$,  if $D(\omega^*) = 1$ and $\widetilde{Y}(\omega^*) =\varnothing$ if $D(\omega^*) = 0$.
Note that since $D\perp(X,Y)$, for any measurable set $A$,
\begin{align*}
P^*(X^* \in A | \widetilde{Y}=\tilde{y})&=	P^*(X^* \in A | \widetilde{Y}=y)\\
	&= \frac{P^*(X^* \in A,D=1,Y^*=y)}
	{P^*(D=1,Y^*=y)} \\
	& = \frac{\pi_N \cdot P(X\in A,Y=y)}{\pi_N\cdot  P(Y=y)}
	= P(X\in A| Y=y), \quad \tilde{y}=y\in\mathcal{Y},\\
P^*(X^* \in A | \widetilde{Y}=\tilde{y})&=	P^*(X^* \in A |\widetilde{Y}=\varnothing)\\
	&= \frac{P^*(X^* \in A,D=0)}{P^*(D=0)}\\
& = \frac{(1-\pi_N)\cdot P(X\in A)}{1-\pi_N}
	= P(X\in A),\quad \tilde{y} = \varnothing.
\end{align*}
\iffalse
\begin{itemize}
	\item  For $\tilde{y} = y \in \mathcal{Y}$,
	\begin{align*}
		P^*(X^* \in A | \widetilde{Y} = y) &= \frac{P^*(X^* \in A, D = 1, Y^* = y)}{P^*(D = 1, Y^* = y)} \\
		&= \frac{\pi_N \cdot P(X \in A, Y = y)}{\pi_N \cdot P(Y = y)} = P(X \in A | Y = y).
	\end{align*}
	\item
	for $\tilde{y} = \star$,
	\begin{align*}
		P^*(X^* \in A | \widetilde{Y} = \star) &= \frac{P^*(X^* \in A, D = 0, Y^* = \star)}{P^*(D = 0, Y^* =  \star)} \\
		& = \frac{(1-\pi_N) \cdot P(X \in A)}{1-\pi_N } = P(X \in A).
	\end{align*}
\end{itemize}
\fi
This confirms the identification.
We hereafter denote $P^*$ simply by $P$, and write $P_{X|\widetilde{Y}}$ for the conditional distribution of $X$ given  $\widetilde{Y}$.

\subsection{Model Specification}
\label{sec:sec_lf}
We represent each label $y$ by its
one-hot encoding $e_y \in \mathbb{R}^K$. Consider the function
$h: \widetilde{\mathcal{Y}} \to \mathbb{R}^{d_h}$ defined by
\begin{equation}
\label{Defh1}
h(\tilde{y}) = b_0 + \Phi_0 e_y \cdot \mathbf{1}\{\tilde{y} = y \in \mathcal{Y}\},
\end{equation}
where $b_0 \in \mathcal{K} \subseteq \mathbb{R}^{d_h}$,
$\Phi_0 \in \mathcal{W} \subseteq \mathbb{R}^{d_h \times K}$, with $\|\Phi_0 e_y\| > 0$ for all
$y \in \mathcal{Y}$.

We assume that the conditional density takes the form
$
p_{X|\widetilde{Y}=\tilde{y}}(x|\tilde{y}) = p(x; h(\tilde{y}))$ for all  $\tilde{y} \in \tilde{\mathcal{Y}}.
$ In particular,
\begin{equation}
\label{eq:additive}
p_X(x) = p(x; b_0), \qquad
p_{X|Y}(x|y) = p(x;b_0 + \Phi_0 e_y), \quad \forall\, y \in \mathcal{Y}.
\end{equation}

This formulation allows abundant unlabeled observations to directly inform the estimation of the shared function $p(\cdot; \cdot)$, while limited labeled pairs identify the specific conditional dependence through $\Phi_0$.

\subsection{Label-Augmented Conditional Diffusion}
\label{sec:ssl_diffusion}

We employ a score-based diffusion model \citep{song2019generative,song2020score}, formulated as a continuous-time stochastic process based on the Ornstein-Uhlenbeck (OU) process \citep{uhlenbeck1930theory}, to learn
$P_{X|\widetilde{Y}=\tilde{y}}$ for all $\tilde{y} \in \widetilde{\mathcal{Y}}$
simultaneously.

Given $\widetilde{Y}=\tilde{y}$,
the forward process stochastic differential equation (SDE) is
\begin{align}
\label{eq:forward_sde}
\mathrm{d} X_t(\tilde{y}) = -\frac{1}{2}\beta_t X_t(\tilde{y})\mathrm{d} t + \sqrt{\beta_t}\mathrm{d} W_t,
\quad\text{with}\quad X_0(\tilde{y}) \sim P_{X|\widetilde{Y}=\tilde{y}}, \ t \in [0, T],
\end{align}
where $\beta_t$ is the time-dependent diffusion coefficient, $W_t$ is a
standard Wiener process. Here and below, $T>0$ is a terminal time  that  can depend on the sample size and diverge to infinity.
Denote the corresponding marginal conditional density function
of $X_t(\tilde{y})=(X_t|\widetilde{Y}=\tilde{y})$ at $t$ by  $p_t(x| \tilde{y}).$ It can be shown that the conditional distribution of $X_t(\tilde{y}) | X_0(\tilde{y})$ is Gaussian with
\begin{align*}
\mathbb{E}[X_t(\tilde{y}) | X_0(\tilde{y})=x_0(\tilde{y})] = \alpha_t x_0(\tilde{y}), \
\mathrm{Cov}(X_t(\tilde{y})|X_0(\tilde{y})=x_0(\tilde{y}) )=\sigma_t^2 I_{d_x},
\end{align*}
where
$\alpha_t = \exp (-\int_0^t \beta_s\,\mathrm{d} s/2)$ and $\sigma_t^2=1-\alpha_t^2.$
So we can represent $X_t$ as
\begin{equation}
\label{vp_xt}
X_t(\tilde{y})= \alpha_t X_0(\tilde{y}) + \sigma_t\epsilon,\ \epsilon \sim N(0, I_{d_x}),
\end{equation}
This expression makes the computation of the forward process (\ref{eq:forward_sde}) straightforward.

The corresponding backward process $(\overline{X}_t(\tilde{y}))_{t\in[0,T]}=(X_{T-t}(\tilde{y}))_{t\in[0,T]}$ is given by \citep{anderson1982}
\begin{align*}
%\label{eq:backward_sde}
d\overline{X}_t(\tilde{y}) = \frac{1}{2}\beta_{T-t} \left\{\overline{X}_t(\tilde{y})
+ 2\nabla_{x_t}  \log p_{T-t}(\overline{X}_t|\tilde{y})\right\}\mathrm{d} t + \sqrt{\beta_{T-t}}
d\overline{W}_t,
\end{align*}
where $\overline{X}_0(\tilde{y})\sim p_{T}(\cdot|\tilde{y})$, $\nabla_{x_t} \log p_t(x_t|\tilde{y})$ is the conditional score function,
and $\overline{W}_t$ is a time-reversed Wiener process.
Let $\overline{Q}_t(\tilde{y})$ denote the distribution of $\overline{X}_t(\tilde{y})$.
%Denote the distribution of $\overline{X}_t(\tilde{y})$ at time $t$ as $\overline{Q}_t(\tilde{y})$.
By the time-reversal
theorem \citep{anderson1982}, $\overline{X}_T(\tilde{y}) \sim
P_{X|\widetilde{Y}=\tilde{y}}$.

 Under the model specification in Section \ref{sec:sec_lf}, the forward diffusion process induces the time-$t$ density
$$p_t(x_t|\tilde{y})=p_t(x_t;h(\tilde{y})) = \int \phi_t(x_t|x) p(x; h(\tilde{y})) dx,$$
where $\phi_t(x_t|x)$ is the Gaussian transition kernel with
$\nabla\log\phi_t(x_t|x) = -(x_t-\alpha_t x)/\sigma_t^2$,  $\alpha_t=\exp\{-\int_{0}^{t}\beta_s \mathrm{d} s/2\}, \ \text{and} \ \sigma_t^2=1-\alpha^2_t$.
We therefore define the true score function as
$s(x_t, h(\tilde{y}), t) := \nabla_{x_t} \log p_t(x_t;h(\tilde{y}))$.
Substituting \eqref{Defh1}, the marginal and conditional score functions are $s(x_t, b_0, t)$ and $s(x_t, b_0+\Phi_0e_y, t)$, respectively.

 We parameterize the score network as $s_{\theta}(x,h_{b,\Phi}(\tilde{y}), t):\mathbb{R}^{d_x} \times \mathbb{R}^{d_h} \times [0,T] \rightarrow \mathbb{R}^{d_x}$, where $s_{\theta}$ is a score network with parameter $\theta$ and $h_{b,\Phi}(\tilde{y})=b+\Phi e_y\cdot \mathbf{1}\{\tilde{y} = y \in \mathcal{Y}\}$ consists of parameter $\Phi \in \mathbb{R}^{d_h\times K}$ and $b\in \mathbb{R}^{d_h}$.

The parameters $(\theta,b, \Phi)$ are estimated by minimizing the denoising score
matching (DSM; \citealt{vincent2011connection,song2020score}) objective $ \mathbb{E}[\ell (X, \tilde{Y}; \theta, b, \Phi) ]$, where
\begin{align*}
\ell (x, \tilde{y}; \theta, b, \Phi) = \int_{\tau}^T \mathbb{E}_{X_t \sim \mathcal{N}
(\alpha_t x, \sigma_t^2 I_{d_x})} \left[\left\|s_{\theta}(X_t, h_{b,\Phi}(\tilde{y}), t)
+ \frac{X_t-\alpha_t x}{\sigma_t^2}\right\|^2\right]\mathrm{d} t.
\end{align*}
Here $\tau$ is an early-stopping time, which can depend on the sample size and converge to zero in our theoretical analysis.

Given the i.i.d. copies $\{(X_i, \widetilde{Y}_i), i=1, \ldots, N\}$, one can minimize
the empirical loss 
\begin{align}
\label{emloss1}
\frac{1}{N}
\sum_{i=1}^N \ell(X_i, \widetilde{Y}_i;  \theta, b, \Phi)
= \frac{N_{\ell}}{N}\cdot \frac{1}{N_{\ell}}
\sum_{i=1}^{N_{\ell}} \ell(X_i, Y_i;  \theta, b, \Phi)+ \frac{N_u}{N}\cdot \frac{1}{N}
\sum_{i=1}^{N_u} \ell(X_i, \varnothing;  \theta, b),
\end{align}
where the unlabeled loss depends only on $(\theta, b)$ since
$h_{b,\Phi}(\varnothing) = b$ does not involve $\Phi$. The empirical
minimizer $(\hat{\theta}, \hat{b}, \hat{\Phi})$ provides the learned
score function $\hat{s}(x, \hat{h}(\tilde{y}), t)
:= s_{\hat{\theta}}(x, \hat{h}_{\hat{b}, \hat{\Phi}}(\tilde{y}), t)$.

Given the learned score function, we generate samples by running the reverse process.
As $p_T(\cdot|\tilde{y})$ is well approximated by $\mathcal{N}(0,I_{d_x})$ when $T$ is large, the approximate backward process is
\begin{align}
\label{eq:approx_backward}
\mathrm{d}\widehat{X}_t(\tilde{y}) = \frac{1}{2}\beta_{T-t}\left\{\widehat{X}_t(\tilde{y})
+ 2\hat{s}(\widehat{X}_t(\tilde{y}), \hat{h}(\tilde{y}), T-t)\right\} \mathrm{d} t
+ \sqrt{\beta_{T-t}}\mathrm{d}\overline{W}_t, t \in [0, T),
\end{align}
with $\widehat{X}_0(\tilde{y})\sim \mathcal{N}(0,I_{d_x})$,
terminated at time $T-\tau$, yielding the output distribution $\widehat{Q}_{T-\tau}(\tilde{y}).$
For conditional generation from $P_{X|Y=y},$  we set $\tilde{y} = y$.

\subsection{Examples of conditional models}
The additive structure~\eqref{eq:additive} encompasses a broad range of distribution families. We illustrate this with the following three representative examples. These examples are not mutually exclusive; for instance, the Gaussian family simultaneously instantiates all three structures. Let $p_Y(\cdot)$ denote the marginal distribution of $Y$.

\subsubsection{Location Family}
\label{ex:location-family}
Consider the location family $p(x|y) = f(x-\mu_y)$, where $\mu_y \in \mathbb{R}^{d_x}$
is the location parameter and $p_Y(y) = 1/K$ for $y \in \{0,\ldots,K-1\}$.

Take $d_h = K+1$ and set $b_0 = (1,0,\ldots,0)^\top \in \mathbb{R}^{K+1}$ and
$\Phi_0=(-\mathbf{1}_K^\top,I_K^\top)^\top\in\mathbb{R}^{(K+1)\times K}$,
where $\mathbf{1}_K$ is the $K$-vector of ones and $I_K$ is the $K\times K$
identity matrix.
\iffalse
\begin{align*}
\Phi_0 = \begin{pmatrix} -1 & -1 & \cdots & -1 \\
1 & 0 & \cdots & 0 \\
0 & 1 & \cdots & 0 \\
\vdots & & \ddots & \vdots \\
0 & 0 & \cdots & 1 \end{pmatrix} \in \mathbb{R}^{(K+1)\times K}.
\end{align*}
\fi
Writing $h = (h_1,\ldots,h_{K+1})^\top$, define
\begin{align*}
p(x;h) = h_1 \cdot \frac{1}{K}\sum_{y=0}^{K-1}f(x-\mu_y)
+ \sum_{y=0}^{K-1}h_{y+2} \cdot f(x-\mu_y).
\end{align*}
Since $b_0 + \Phi_0 e_y = (0,\ldots,0,1,0,\ldots,0)^\top$ with $1$ in position $y+2$,
we have $p(x;b_0) = \frac{1}{K}\sum_{y=0}^{K-1}f(x-\mu_y) = p_X(x)$
and $p(x;b_0+\Phi_0 e_y) = f(x-\mu_y) = p(x|y)$.

\subsubsection{Exponential family}
\label{ex:exponential-family}
Consider the exponential family
$p(x|y) = \exp\bigl(\langle \eta_y,\,T(x)\rangle - A(\eta_y)\bigr)$,
where $\eta_y \in \mathbb{R}^{d_\eta}$ is the natural parameter and $T(\cdot)$ is a fixed sufficient statistic. The marginal $p_X(x)$ is a mixture of exponential family members.
Similarly, we can construct $p(x;h)$ satisfying~\eqref{eq:additive} by the embedding procedure used for the location family.

\subsubsection{Latent variable model}
\label{ex:latent-variable-model}
Consider two types of latent variable models.
(i) \textit{$h$ in the prior.}
Let $p(x|y) = \int p(x|z)p(z|y)\mathrm{d}z$.
Define the parameterized model $p(x;h) = \int p(x|z)p(z;h)\mathrm{d}z$,
and set $p(z;b_0+\Phi_0 e_y) = p(z|y)$ and
$p(z;b_0) = K^{-1}\sum_{y=0}^{K-1}p(z|y) = p(z)$.
Then \eqref{eq:additive} is directly satisfied without additional embedding.
(ii) \textit{$h$ in the decoder.}
Let $p(x|y) = \int p(x|z,\eta_y)p(z)\mathrm{d}z$,
where $\eta_y$ is the condition-specific parameter.
The marginal distribution simplifies to
$p_X(x) = K^{-1}\sum_{y=0}^{K-1}\int p(x|z,\eta_y)p(z)\mathrm{d}z$.
We can apply the same embedding method used for the location family.

\begin{remark}
The construction of $b_0$ and $\Phi_0$ satisfying \eqref{eq:additive} is not
unique.
For instance, in the location family example above, taking $d_h = K+1$
gives the construction in Section~\ref{ex:location-family}; taking $d_h = K+2$
with $b_0 = (1,0,\ldots,0)^\top \in \mathbb{R}^{K+2}$ and $\Phi_0$
augmented by an additional zero row also satisfies \eqref{eq:additive}. More
generally, any $d_h \geq K+1$ admits a valid construction.
\end{remark}

\section{Convergence Analysis}
\label{sec:convergence_analysis}

This section establishes finite-sample convergence rates for the LACD estimator in Wasserstein-1 ($W_1$) and total variation (TV) distances.
We begin by introducing the necessary notations and function classes, followed by the assumptions and main results.

\medskip
\noindent \textbf{Notations and Metrics.}
Let $\|\cdot\|$ be the Euclidean norm.
For two probability distributions $\nu_1$ and $\nu_2$ on $\mathbb{R}^{d}$, define
\begin{align*}
W_1(\nu_1, \nu_2) = \inf_{\gamma \in \Pi(\nu_1,\nu_2)}
\mathbb{E}_{(Z, Z') \sim \gamma}[\|Z - Z'\|],\quad \text{and} \quad
\mathrm{TV}(\nu_1,\nu_2) = \sup_A |\nu_1(A)-\nu_2(A)|,
\end{align*}
where $\Pi(\nu_1, \nu_2)$ is the set of all joint distributions whose marginals
are $\nu_1$ and $\nu_2$ respectively, and the supremum is taken over all
measurable sets $A \subset \mathbb{R}^{d}$.

Let $\|v\|_\infty = \max_j |v_j|$ for $v \in \mathbb{R}^d$; $\|B\|_0 = \sum_{i,j} \mathbf{1}_{\{B_{ij} \neq 0\}}$ and $\|B\|_\infty = \max_{i,j} |B_{ij}|$ for matrix $B$; and $\|f\|_\infty = \sup_{z}\|f(z)\|$ for measurable map $f(\cdot)$.  %$\|f\|_{L_2(P)} = (\mathbb{E}_P[\|f(Z)\|^2])^{1/2}$ for measurable map $f: \mathbb{R}^{d_x} \to \mathbb{R}^k$ and $Z \sim P$.
We write $a_n \lesssim b_n$ (or $a_n = O(b_n)$) if $a_n \leq Cb_n$ for some constant $C > 0$ independent of $n$, and write $a_n \gtrsim b_n$ if $b_n \lesssim a_n$. We denote $a_n \asymp b_n$ (or $a_n = \Theta(b_n)$) if $a_n \lesssim b_n \lesssim a_n$.
Based on these, $\widetilde{O}(b_n)$ (resp. $\widetilde{\Theta}(b_n)$) represents $O(b_n \log^k n)$ (resp. $\Theta(b_n \log^k n)$) for some $k \geq 0$. 
For $a, b \in \mathbb{R}$, we write $a \vee b = \max\{a,b\}$
and $a \wedge b = \min\{a,b\}$.

\medskip
\noindent \textbf{Neural Network Classes.}  The score network $s_\theta$
introduced in Section~\ref{sec:sec_lf} is instantiated as ReLU neural network.
We formalize the neural network classes below.
\begin{definition}[ReLU Neural Network Function Class]
\label{def:neural-network-class}
Consider the functions $f: \mathbb{R}^{d_{\text{in}}} \to \mathbb{R}^{d_{\text{out}}}$ with the compositional structure
\begin{align*}
f(z) = (A_L \sigma(\cdot) + b_L) \circ \cdots \circ (A_2 \sigma(\cdot) + b_2) \circ (A_1 z + b_1),
\end{align*}
where $\sigma(x) = \max\{0, x\}$ is the ReLU activation applied element-wise, $A_i \in \mathbb{R}^{d_i \times d_{i-1}}$ and $b_i \in \mathbb{R}^{d_i}$ for $i = 1, \ldots, L$ with $d_0 = d_{\text{in}}$ and $d_L = d_{\text{out}}$, subject to the constraints $\max_{i=0,\ldots,L} d_i \leq W$, $\sup_{z \in \mathbb{R}^{d_{\text{in}}}} \|f(z)\|_\infty \leq S$, $\max_{i=1,\ldots,L} (\|A_i\|_\infty \vee \|b_i\|_\infty) \leq D$, and $\sum_{i=1}^L (\|A_i\|_0 + \|b_i\|_0) \leq R$.
Denote the set of such functions by
$\Phi(d_{\text{in}}, d_{\text{out}}, L, W, S, D, R).$
\end{definition}

\subsection{Error Decomposition}
\label{sec:error_decomp}
The analysis relies on the following function classes and network architectures.

We write $	\mathcal{F}^\beta(U, \mathbb{R}^{d_{\mathrm{o}}}, B) =
\{ f: U \to \mathbb{R}^{d_{\mathrm{o}}} \mid \|f\|_{\mathcal{F}^\beta} \leq B \}$
as  the H\"{o}lder function class.
% with formal definition deferred to the Supplementary Materials.

The score function estimator is assumed to belong to a time-varying
neural network class $\Gamma$, as specified below.
\begin{align*}
\Gamma = \Big\{ & s_\theta(x, h_{b,\Phi}(\tilde{y}), t) = \sum_{i=1}^{\mathcal{J}}
s_{\theta,i}(x, h_{b,\Phi}(\tilde{y}), t) \cdot \mathbf{1}(t_{i-1} \leq t < t_i) :\\
& s_{\theta,i} \in \Phi(d_x+d_h+1, d_x, L_i, W_i, S_i, D_i, R_i)
\text{ for } i \in [J], \\
& h_{b,\Phi}(\tilde{y}) = b + \Phi e_y \cdot \mathbf{1}\{\tilde{y} = y \in \mathcal{Y}\},\Phi \in \mathcal{W},
\ b \in \mathcal{K} \Big\},
\end{align*}
where $\tau = t_0 < t_1 < \cdots < t_{\mathcal{J}} = T$ with $t_{i+1}/t_i = 2$ for all $i \in \{0, 1, \ldots, \mathcal{J}-1\}$ and $\tau = 2^{-\mathcal{J}} T$.

The key to establishing convergence rates is to control the score estimation error.
By a SDE error propagation argument (see SM), combined with Jensen's inequality,
\begin{align*}
\mathbb{E}_{\mathcal{D}}\bigl[\mathbb{E}_{Y}[
W_1(P_{X|Y}, \widehat{Q}_{T-\tau}(Y))]\bigr]
&\lesssim e^{-T} + \sqrt{\tau}
+ \sum_{i=0}^{\mathcal{J}-1}\sqrt{(t_i\log N)\wedge 1}
\cdot\sqrt{\mathbb{E}_{\mathcal{D}}[
\mathcal{L}_{\mathrm{c},i}(\hat\theta,\hat b,\hat\Phi)]},\\[6pt]
\mathbb{E}_{\mathcal{D}}\bigl[\mathbb{E}_{Y}[
\mathrm{TV}(P_{X|Y}, \widehat{Q}_{T-\tau}(Y))]\bigr]
&\lesssim e^{-T}
+ \mathbb{E}_{\mathcal{D}}\bigl[\mathbb{E}_{Y}[
\mathrm{TV}(\overline{Q}_{T}(Y), \overline{Q}_{T-\tau}(Y))]\bigr]+ \sum_{i=0}^{\mathcal{J}-1}\sqrt{\mathbb{E}_{\mathcal{D}}[
\mathcal{L}_{\mathrm{c},i}(\hat\theta,\hat b,\hat\Phi)]},
\end{align*}
where
\begin{align}
\label{eq:con_localized_loss}
\mathcal{L}_{\mathrm{c},i}(\theta, b, \Phi)
:= \int_{t_i}^{t_{i+1}}
\mathbb{E}_{X_t, Y}\bigl[\bigl\|s_{\theta}(X_t, b+\Phi e_Y, t)
- \nabla_{x_t} \log p_{t}(X_t|Y)\bigr\|^2\bigr]\,\mathrm{d}t
\end{align}
is the conditional score matching loss over $[t_i, t_{i+1}]$.
Second, we bound $\mathbb{E}_\mathcal{D}[
\mathcal{L}_{\mathrm{c},i}(\hat\theta,\hat b,\hat{\Phi})]$
by decomposing it into statistical and approximation errors,
and balance these errors. % optimally, yielding Theorem~\ref{thm:distributional_convergence_d}.

Denote the Fisher divergence as
\begin{align}
\label{eq:fisher_loss}
\mathcal{L}(\theta, b, \Phi)=&\int_\tau^T \mathbb{E}_{X_t, \widetilde{Y}}[\|s_\theta(X_t, h_{b,\Phi}(\widetilde{Y}), t) - \nabla_{x_t} \log p_t(X_t|\widetilde{Y})\|^2]\mathrm{d}t\\
=& \pi_N  \mathcal{L}_{\text{c}}(\theta, b, \Phi) +(1-\pi_N) \mathcal{L}_{\text{m}}(\theta, b), \nonumber
\end{align}
where $ \mathcal{L}_{\text{c}}(\theta, b, \Phi) =\sum_{i=1}^{\mathcal{J}} \mathcal{L}_{\text{c},i}(\theta, b,\Phi)$, $ \mathcal{L}_{\text{m}}(\theta, b) =\sum_{i=1}^{\mathcal{J}} \mathcal{L}_{\text{m},i}(\theta, b)$.
Here
\begin{align*}
\mathcal{L}_{\text{m},i}(\theta, b) = \int_{t_i}^{t_{i+1}}
\mathbb{E}_{X_t}\left[\left\|s_{\theta}(X_t, b, t)
- \nabla_{x_t} \log p_{t}(X_t)\right\|^2\right]\mathrm{d}t.
\end{align*}
Recall that $\hat{s}(x, \hat{h}(\tilde{y}), t)
= s_{\hat{\theta}}(x, \hat{h}_{\hat{b}, \hat{\Phi}}(\tilde{y}), t)$ and
the parameters $(\hat{\theta}, \hat{b}, \hat{\Phi})$ are estimated by minimizing empirical DSM objective \eqref{emloss1}.
By the equivalence of score matching objectives
\citep{vincent2011connection}, minimizing the DSM objective is equivalent to minimizing the Fisher divergence \eqref{eq:fisher_loss}.
Therefore, our first goal is to bound $$\mathbb{E}_{\mathcal{D}}[\mathcal{L}_{\text{c},i}
(\hat{\theta}, \hat{b}, \hat{\varphi})],\quad \text{ for each }  i \in [\mathcal{J}].$$

Define $(\bar{\theta}, \bar{b}, \bar{\Phi}) \in \arg\min_{\Gamma}
\mathcal{L}(\theta, b, \Phi)$.
For any shared parameter $(\theta, b)$, define the profile minimizer
$$\bar{\Phi}_{\theta,b} \in \arg\min_{\Phi \in \mathcal{W}}
\mathcal{L}_{\mathrm{c}}(\theta, b, \Phi).$$
Denote $\Delta_{\mathrm{c}}(\theta,b) = \mathcal{L}_{\mathrm{c}}(\theta, b,
\bar{\Phi}_{\theta,b})$ and $\Delta_{\mathrm{m}}(\theta,b) =
\mathcal{L}_{\mathrm{m}}(\theta,b)$.
Let $\mathcal{H} := \{h(y) : y \in \mathcal{Y}\}$ denote the range
of $h(\cdot)$ over $\mathcal{Y}$.

We now state the assumptions for the theoretical analysis.
\begin{assumption}[Compact Support]
\label{ass:compact-support-d}
The sets
$\mathcal{X}$, $\mathcal{W}$ and $\mathcal{K}$ are compact.
\end{assumption}

\begin{assumption}[Density Regularity]
\label{ass:sec_holder_d}
\noindent
\begin{itemize}
\item[(H1)]
$p_X(x) \in \mathcal{F}^{\beta_x}(\mathcal{X}, \mathbb{R}, B_{xh})$
and $p(x;h) \in \mathcal{F}^{\beta_x}(\mathcal{X}, \mathbb{R}, B_{xh})$
uniformly over $h \in \mathcal{H}$, for some $B_{xh} > 0$
and $\beta_x > 0$.
\item[(H2)] There exist $0 < p_{\min} \leq p_{\max} < \infty$ such that
$p_{\min} \leq p(x;h(\tilde{y})) \leq p_{\max}$ for all $x \in \mathcal{X}$
and $\tilde{y} \in \widetilde{\mathcal{Y}}$.
\end{itemize}
\end{assumption}

\begin{assumption}[Structural Consistency]
\label{ass:structural_consistency}
There exists a constant $c_1 > 0$ such that for all $(\theta, b)$:
$|\Delta_{\text{c}}(\theta, b) - \Delta_{\text{c}}(\bar{\theta},\bar{b})| \leq c_1 |\Delta_{\text{m}}(\theta, b) - \Delta_{\text{m}}(\bar{\theta},\bar{b})|$.
\end{assumption}

\begin{assumption}[Diffusion Schedule]
\label{ass:schedule}	
The schedule $\beta_t$ is continuous, non-decreasing, and bounded by some $\underline{\beta}, \bar{\beta}> 0$ such that for any $t \in [0,T]$, $\underline{\beta} \leq \beta_t \leq \bar{\beta}$.
\end{assumption}

Assumption~\ref{ass:compact-support-d} is a standard regularity condition
ensuring the boundedness of the support.
Assumption~\ref{ass:sec_holder_d} imposes smoothness on the data-generating
process.
%: both the marginal density and the conditional density are assumed to belong to $\mathcal{F}^{\beta_x}$.
Assumption~\ref{ass:structural_consistency} formalizes task alignment by
requiring the conditional performance gap $\Delta_{\mathrm{c}}(\theta,b)$
to be controlled by the marginal discrepancy $\Delta_{\mathrm{m}}(\theta,b)$,
ensuring that a near-optimal shared parameter $(\theta,b)$ provides a
sufficient foundation for conditional distribution estimation.
This condition parallels structural assumptions in multi-task and transfer
learning \citep{tripuraneni2020theory, chen2025enhancingcausaleffectestimation,
tian2025enhancingaccuracygenerativemodels}.
Assumption~\ref{ass:schedule} is standard in the diffusion model literature.
Several common scheduling options, including constant schedule, linear
schedule \citep{ho2020denoising}, and cosine schedule
\citep{nichol2021improved}, satisfy this condition.

We decompose $\mathcal{L}_{\mathrm{c},i}(\hat{\theta}, \hat{b}, \hat{\Phi})$
into three components:
\begin{align}
\label{eq:error_decomp}
\mathcal{L}_{\mathrm{c},i}(\hat{\theta}, \hat{b}, \hat{\Phi}) =
\mathcal{E}_{\mathrm{emb},i} + \mathcal{E}_{\mathrm{shared},i} + \mathcal{E}_{\mathrm{approx},i},
\end{align}
where
\begin{align*}
\mathcal{E}_{\mathrm{emb},i} &:= \mathcal{L}_{\mathrm{c},i}(\hat{\theta}, \hat{b}, \hat{\Phi}) - \mathcal{L}_{\mathrm{c},i}(\hat{\theta}, \hat{b}, \bar{\Phi}_{\hat{\theta},\hat{b}}), \\
\mathcal{E}_{\mathrm{shared},i} &:= \mathcal{L}_{\mathrm{c},i}(\hat{\theta}, \hat{b}, \bar{\Phi}_{\hat{\theta},\hat{b}}) - \mathcal{L}_{\mathrm{c},i}(\bar{\theta}, \bar{b}, \bar{\Phi}_{\bar{\theta},\bar{b}}), \\
\mathcal{E}_{\mathrm{approx},i} &:= \mathcal{L}_{\mathrm{c},i}(\bar{\theta}, \bar{b}, \bar{\Phi}_{\bar{\theta},\bar{b}}).
\end{align*}
Here $\mathcal{E}_{\mathrm{emb},i}$ captures the estimation error of the
embedding matrix $\Phi$ given fixed shared parameters $(\hat{\theta}, \hat{b})$;
$\mathcal{E}_{\mathrm{shared},i}$ captures the estimation error of the shared
parameters $(\theta, b)$; and $\mathcal{E}_{\mathrm{approx},i}$ is the
approximation error measuring the best achievable loss within $\Gamma$.

Denote the statistical complexity terms as
\begin{align*}
\mathfrak{C}_{\Phi} := \frac{d_h K \log N}{N_{\ell}}, \quad \text{and} \quad
\mathfrak{C}_{\theta,b,i} := (\log N)^2\,\frac{R_i L_i \log(L_i(D_i\vee 1)W_i N)}{N_u}.
\end{align*}

The following lemma bounds $\mathcal{E}_{\text{emb},i}$.

\begin{lemma}
\label{lem:estimation_error_emb}
Under Assumptions \ref{ass:compact-support-d}--\ref{ass:structural_consistency},
with probability at least $1-1/N_{\ell}$, for all $i \in [\mathcal{J}]$,
\begin{align}
\label{eq:estimation_error_emb}
\mathcal{E}_{\mathrm{emb},i} \lesssim \mathfrak{C}_{\Phi} +
\mathcal{E}_{\mathrm{approx},i} +
|\mathcal{L}_{\mathrm{m},i}(\hat{\theta}, \hat{b}) -
\mathcal{L}_{\mathrm{m},i}(\bar{\theta}, \bar{b})|.
\end{align}
\end{lemma}

By Assumption~\ref{ass:structural_consistency}, $\mathcal{E}_{\text{shared},i}$ satisfies
$|\mathcal{E}_{\text{shared},i}| \leq c_1|\mathcal{L}_{\text{m},i}(\hat{\theta}, \hat{b}) - \mathcal{L}_{\text{m},i}(\bar{\theta}, \bar{b})|$.
The following lemma bounds $\mathcal{E}_{\text{shared},i}$.

\begin{lemma}
\label{lem:estimation_error_shared}
Under Assumptions \ref{ass:compact-support-d}--\ref{ass:structural_consistency},
with probability at least $1-1/N_u$, for all $i \in [\mathcal{J}]$,
\begin{align}
\label{eq:estimation_error_shared}
\mathcal{E}_{\text{shared},i}\lesssim
\mathfrak{C}_{\theta,b,i} + \mathcal{L}_{\text{m},i}(\bar{\theta}, \bar{b}).
\end{align}
\end{lemma}

Lemma \ref{lem:total_approx_error_d} bounds the approximation errors $\mathcal{E}_{\text{approx},i}$
and $\mathcal{L}_{\text{m},i}(\bar{\theta}, \bar{b})$, which measure the best
approximation of the true score achievable within $\Gamma$.

\begin{lemma}
\label{lem:total_approx_error_d}
Under Assumptions \ref{ass:compact-support-d}--\ref{ass:structural_consistency},
let $\delta > 0$ be a target precision, and let
$t^* = \delta^{2/\beta_x}/\log N$,
$\Lambda = \Theta(\log N \cdot \log\delta^{-1})$,
$\Lambda_\tau = \Theta(\log N \cdot \log\tau^{-1})$.
There exists $\bar{s} \in \Gamma$ with score network parameters
for each $i \in [\mathcal{J}]$:
\begin{itemize}
\item[(i)] If $t^* \leq t_i \leq T$:
$L_i = \Theta(\Lambda^2 + \log^2 N)$,
$W_i = \widetilde{\Theta}(t_i^{-d_x/2}\Lambda^3)$,
$D_i = \exp(\Theta(\Lambda^2)) \vee O(\log N \cdot t_i^{-1})
\vee \exp(\Theta(\log N))$,
$S_i = \Theta(\sqrt{\log N/(t_i \wedge 1)})$,
$R_i = \widetilde{\Theta}(t_i^{-d_x/2}\Lambda^4)$,
\item[(ii)] If $\tau \leq t_i \leq t^*$:
$L_i = \Theta(\Lambda_\tau^2 + \log^2 N)$,
$W_i = \widetilde{\Theta}(\delta^{-d_x/\beta_x}\Lambda_\tau^3)$,
$D_i = \exp(\Theta(\Lambda_\tau^2)) \vee O(\log N \cdot \delta^{-2/\beta_x})
\vee \exp(\Theta(\log N))$,
$S_i = \Theta(\sqrt{\log N/t_i})$,
$R_i = \widetilde{\Theta}(\delta^{-d_x/\beta_x}\Lambda_\tau^4)$,
\end{itemize}
such that
\begin{align}
\label{eq:approx_bound_d}
\mathcal{E}_{\mathrm{approx},i} = \widetilde{O}(\delta^2),
\quad
\mathcal{L}_{\mathrm{m},i}(\bar{\theta}, \bar{b}) = \widetilde{O}(\delta^2).
\end{align}
\end{lemma}

Using the error decomposition
\eqref{eq:error_decomp}, \eqref{eq:estimation_error_emb},
\eqref{eq:estimation_error_shared}, and \eqref{eq:approx_bound_d},
and balancing the resulting statistical and approximation terms, we control
the two score-error sums appearing in the SDE error-propagation inequalities:
$\sum_{i=0}^{\mathcal{J}-1}\sqrt{(t_i\log N)\wedge 1}
\cdot\sqrt{\mathbb{E}_\mathcal{D}[\mathcal{L}_{\mathrm{c},i}]}$
and
$\sum_{i=0}^{\mathcal{J}-1}
\sqrt{\mathbb{E}_\mathcal{D}[\mathcal{L}_{\mathrm{c},i}]}$.
Recall that $\widehat Q_{T-\tau}(\tilde y)$ denotes the output distribution
of the practical backward process in~\eqref{eq:approx_backward}.
	For conditional generation, we write the resulting estimator as
	\(\widehat P_{X|Y}\), with
	\(\widehat P_{X|Y}(\cdot\mid y)=\widehat Q_{T-\tau}(y)\) for
	\(y\in\mathcal Y\). The following theorem gives its convergence rates.

\begin{theorem}
\label{thm:distributional_convergence_d}
Suppose Assumptions~\ref{ass:compact-support-d}--\ref{ass:schedule} hold.
Let $T = \Theta(\log N)$ and
$\tau = N_u^{-2(1+1/\beta_x)/(2+d_x/\beta_x)}$, there exists
$\hat{s} \in \Gamma$ with network configuration
\begin{align}
L_i &= \Theta(\log^4 N), \quad
D_i = \exp\bigl(\Theta(\log^4 N)\bigr), \quad
S_i = \Theta(\sqrt{\log N/(t_i \wedge 1)}), \notag\\
W_i &= R_i = \widetilde{\Theta}(
t_i^{-d_x/2} \wedge N_u^{(d_x/\beta_x)/(2+d_x/\beta_x)}).
\label{eq:network_sizes_d}
\end{align}
Then
\begin{align*}
\mathbb{E}_{\mathcal{D}}[\mathbb{E}_{Y}[
W_1(P_{X|Y}, \widehat{P}_{X|Y})]]
&= \widetilde{O}\!\left(N_{\ell}^{-\frac{1}{2}}
+ N_u^{-\frac{1}{2}}
+ N_u^{-\frac{1+\beta_x}{2\beta_x+d_x}}
\right),\\
\mathbb{E}_{\mathcal{D}}[\mathbb{E}_{Y}[
\mathrm{TV}(P_{X|Y}, \widehat{P}_{X|Y})]]
&= \widetilde{O}\!\left(
N_{\ell}^{-\frac{1}{2}}
+ N_u^{-\frac{\beta_x}{2\beta_x+d_x}}
\right).
\end{align*}
\end{theorem}

If we only use labeled data, we have the following results.

\begin{corollary}[Supervised counterpart]
\label{cor:supervised_counterpart_d}
Under Assumptions~\ref{ass:compact-support-d}--\ref{ass:schedule},
with $T = \Theta(\log N_{\ell})$ and
$\tau = N_{\ell}^{-2(1+1/\beta_x)/(2+d_x/\beta_x)}$, there exists
$\hat{s}^{\mathrm{sup}} \in \Gamma$ with network configuration
\begin{align}
L_i &= \Theta(\log^4 N_{\ell}), \quad
D_i = \exp\bigl(\Theta(\log^4 N_{\ell})\bigr), \quad
S_i = \Theta(\sqrt{\log N_{\ell}/(t_i \wedge 1)}), \notag\\
W_i &= R_i = \widetilde{\Theta}(
t_i^{-d_x/2}
\wedge
N_{\ell}^{(d_x/\beta_x)/(2+d_x/\beta_x)}
).
\label{eq:network_sizes_d_sup}
\end{align}
Let  {$\widehat{P}^{\mathrm{sup}}_{X|Y}$}
denote the corresponding
output conditional distribution of the reverse process. Then
\begin{align*}
\mathbb{E}_{\mathcal{D}_l}[\mathbb{E}_{Y}[
W_1(P_{X|Y}, \widehat{P}^{\mathrm{sup}}_{X|Y}
)]]
&= \widetilde{O}\!\left(
N_{\ell}^{-\frac{1}{2}}
+ N_{\ell}^{-\frac{1+\beta_x}{2\beta_x+d_x}}
\right),\\
\mathbb{E}_{\mathcal{D}_l}[\mathbb{E}_{Y}[
\mathrm{TV}(P_{X|Y}, \widehat{P}^{\mathrm{sup}}_{X|Y}
)]]
&= \widetilde{O}\!\left(
N_{\ell}^{-\frac{1}{2}}
+ N_{\ell}^{-\frac{\beta_x}{2\beta_x+d_x}}
\right).
\end{align*}
\end{corollary}

To quantify the gain from unlabeled data, we compare the convergence rates
of LACD with those of its supervised counterpart. Table~\ref{tab:rate_comparison_dominant} summarizes the
dominant rates.
\begin{table}[H]
\centering
\begin{tabular}{l|c|c}
\hline
Metric/regime & Supervised counterpart & LACD \\ \hline
$W_1$ & $N_{\ell}^{-\frac12}+N_{\ell}^{-\frac{1+\beta_x}{2\beta_x+d_x}}$ &
$N_{\ell}^{-\frac12} +N_u^{-\frac12} +N_u^{-\frac{1+\beta_x}{2\beta_x+d_x}}$
\\ \hline
TV
&
$N_{\ell}^{-\frac12}+N_{\ell}^{-\frac{\beta_x}{2\beta_x+d_x}}$
&
$N_{\ell}^{-\frac12}+N_u^{-\frac{\beta_x}{2\beta_x+d_x}}$
\\ \hline
\end{tabular}\\

\medskip
\captionsetup{font=normalsize}
\caption{Dominant convergence rates for LACD and its supervised counterpart, with logarithmic factors suppressed.}
\label{tab:rate_comparison_dominant}
\end{table}

	When $d_x>2$, the supervised $W_1$ rate is dominated by
	$N_{\ell}^{-(1+\beta_x)/(2\beta_x+d_x)}$. Under $ N_{\ell}\lesssim N_u$, we have
	$N_u^{-(1+\beta_x)/(2\beta_x+d_x)}
	\lesssim N_{\ell}^{-(1+\beta_x)/(2\beta_x+d_x)}$. Moreover, for $d_x>2$,
	$N_u^{-1/2}\lesssim N_{\ell}^{-1/2}
	\lesssim N_{\ell}^{-(1+\beta_x)/(2\beta_x+d_x)}$. Hence the convergence rate of
	LACD is no slower than that of the supervised counterpart.
	If further $N_{\ell}/N_u\to0$, then all three terms in the LACD $W_1$ bound are
	of smaller order than the supervised dominant term, and the convergence rate
	of LACD is strictly faster.
	
When $d_x\le2$, the supervised $W_1$ rate is dominated by $N_{\ell}^{-1/2}$.
Under $ N_{\ell}\lesssim N_u$, we have $N_u^{-1/2}\lesssim N_{\ell}^{-1/2}$. Moreover, since $d_x\le2$, $N_u^{-(1+\beta_x)/(2\beta_x+d_x)} \lesssim N_u^{-1/2}\lesssim N_{\ell}^{-1/2}$.
Hence the convergence rate of LACD is no slower than that of the supervised counterpart.

For the TV distance, the convergence rate of supervised counterpart is dominated by
$N_{\ell}^{-\beta_x/(2\beta_x+d_x)}$. Using similar arguments, the convergence rate of
LACD is no slower than that of the supervised counterpart when
$ N_{\ell}\lesssim N_u$, and is strictly faster when $N_{\ell}/N_u\to0$.
	
The above results provide theoretical support for the benefits of using unlabeled data in our proposed method. We also extend our results to the case where $Y$ is continuous.

\section{Numerical Experiments}
\label{sec:Numerical_Experiments}

We evaluate the proposed LACD method through simulation studies on
synthetic datasets (Section~\ref{sec:simulation}), image benchmarks
(Sections~\ref{sec:cifar10} and~\ref{sec:intel6}),
and a tabular EEG benchmark (Section~\ref{sec:beed}).
The general algorithm is presented in Section~\ref{sec:implementation}.
% with dataset-specific implementation details provided in the Supplementary Materials.

\subsection{Implementation}
\label{sec:implementation}

Algorithm~\ref{alg:LACD} describes the LACD procedure used
in simulation and tabular experiments. For image experiments, we adopt
the EDM framework~\citep{karras2022elucidating}; the label augmentation
scheme and score estimation objective remain identical, while the sampler
follows the EDM implementation.

\begin{algorithm}[htp!]
\caption{{\it Label-Augmented Conditional Diffusion (LACD).}}
\label{alg:LACD}
\begin{algorithmic}[1]
\State \textbf{Input:} $\mathcal{D}=\{(X_i,\widetilde{Y}_i)\}_{i=1}^{N}$:
training data.
\State \textbf{Input:} $\beta_t$: diffusion schedule, $T$: terminal time,
$\tau$: early-stopping time,
$\{\zeta_k\}_{k=0}^{K}$: stepsizes with
$\sum_{k=0}^{K}\zeta_k = T-\tau$,
and $r_k = \sum_{j=0}^{k-1}\zeta_j$ with $r_0 = 0$.
\State \textbf{Input:} Target label $y \in \mathcal{Y}$: the condition
for generating new samples.

\State \textbf{Score Function Estimation:} Minimize the empirical DSM
objective:
\[
(\hat{\theta}, \hat{b}, \hat{\Phi})
\;\in\;
\arg\min_{\theta,\, b,\, \Phi}
\;\frac{1}{N}\sum_{i=1}^{N} \ell(X_i, \widetilde{Y}_i;\, \theta, b, \Phi),
\]
yielding the learned score function
$\hat{s}(\cdot\,,\hat{h}(\tilde{y}),\cdot)
:= s_{\hat\theta}(\cdot\,,\hat{h}_{\hat{b},\hat{\Phi}}(\tilde{y}),\cdot)$.

\State \textbf{Sampling Procedure:}
\State \textbf{Initialize:} $\breve{X}_{r_0}(y) = Z \sim \mathcal{N}(0,I_{d_x})$.
\For{$k = 0, \ldots, K-1$}
\State Set $t_{k+1} = \sum_{j=0}^{k} \zeta_j$.
\State
Update:
\[
\breve{X}_{r_{k+1}}(y) = \breve{X}_{r_k}(y)
+ \gamma_{1,k}\!\left\{\breve{X}_{r_k}(y)
+ 2\hat{s}\!\left(\breve{X}_{r_k}(y),\hat{h}(y),T-r_k\right)\right\}
+ \sqrt{\gamma_{2,k}}\,Z_{k},
\]
where $\gamma_{1,k} = \exp\!\left\{\int_{T-r_{k+1}}^{T-r_k}\!\beta_s\,ds/2\right\}-1$,
$\;\gamma_{2,k} = \exp\!\left\{\int_{T-r_{k+1}}^{T-r_k}\!\beta_s\,ds\right\}-1$.
\EndFor
\State \Return $\breve{X}_{r_K}(y)$.
\end{algorithmic}
\end{algorithm}

\subsection{Simulation Studies}
\label{sec:simulation}

We consider nine synthetic two-dimensional datasets
\citep{grathwohl2018scalable,pmlr-v145-22a} with covariates
$X \in \mathbb{R}^2$ and discrete labels
$Y \in \{0,1,\ldots,K-1\}$.
For all nine datasets, we fix $N_{\ell} = 200$ while varying
$N_u \in \{0,200,\ldots,1{,}000,2{,}000\}$.
Generation quality is evaluated by generating $n_{g,y} = 2{,}000$ samples
per class and computing three class-conditional metrics: Total Variation
(TV), Maximum Mean Discrepancy (MMD), and Wasserstein-1 ($W_1$)
distance, each averaged across classes as
$\text{Metric} = K^{-1}\sum_{y \in \mathcal{Y}}\text{Metric}(y)$,
For each configuration, we report the best value attained during training.
The metrics are defined as follows:
\begin{itemize}
	\item[(a)] Total Variation distance $\text{TV}(y)$:
	$\text{TV}(y) = \int |p_{\text{r}}(x|y) - p_{\text{g}}(x|y)| \, dx/2$,
	where $p_{\text{r}}(x|y)$ and $p_{\text{g}}(x|y)$ are estimated via
	two-dimensional Gaussian kernel density estimation with numerical integration.
	\item[(b)] Maximum Mean Discrepancy $\text{MMD}(y)$:
	$\text{MMD}^2(y) = K_{rr} + K_{gg} - 2K_{rg}$, where
	$K_{rr} = \tfrac{1}{n(n-1)} \sum_{i \neq j} k(x_{r,i}^{(y)}, x_{r,j}^{(y)})$,
	$K_{gg} = \tfrac{1}{m(m-1)} \sum_{i \neq j} k(x_{g,i}^{(y)}, x_{g,j}^{(y)})$, and
	$K_{rg} = \tfrac{1}{nm} \sum_{i,j} k(x_{r,i}^{(y)}, x_{g,j}^{(y)})$.
	Here, $k(\cdot, \cdot)$ is a radial basis function (RBF) kernel, and
	$\{x_{r,i}^{(y)}\}_{i=1}^n$, $\{x_{g,j}^{(y)}\}_{j=1}^m$ denote real and
	generated samples for class $y$, respectively.
	\item[(c)] Wasserstein-1 distance $W_1(y)$:
	$W_1(y) = \inf_{\gamma \in \Pi(p_{\text{r}}(x|y),\, p_{\text{g}}(x|y))}
	\mathbb{E}_{(x,z)\sim\gamma}[\|x-z\|]$,
	where $\Pi(\cdot,\cdot)$ denotes the set of couplings. We use the sliced
	Wasserstein approximation~\citep{rabin2011wasserstein} with random projections to one-dimensional spaces.
\end{itemize}

 Here we report three representative examples, \texttt{large\_4gaussians},
\texttt{rings}, and \texttt{2spirals}, which cover well-separated multimodal
clusters, concentric nonlinear structures, and intertwined curved manifolds,
respectively. Complete results for all nine synthetic datasets, closed-form
conditional densities.
% and implementation details are provided in the Supplementary Materials.

\begin{figure}[ht]
	\centering
	\includegraphics[width=0.95\textwidth]{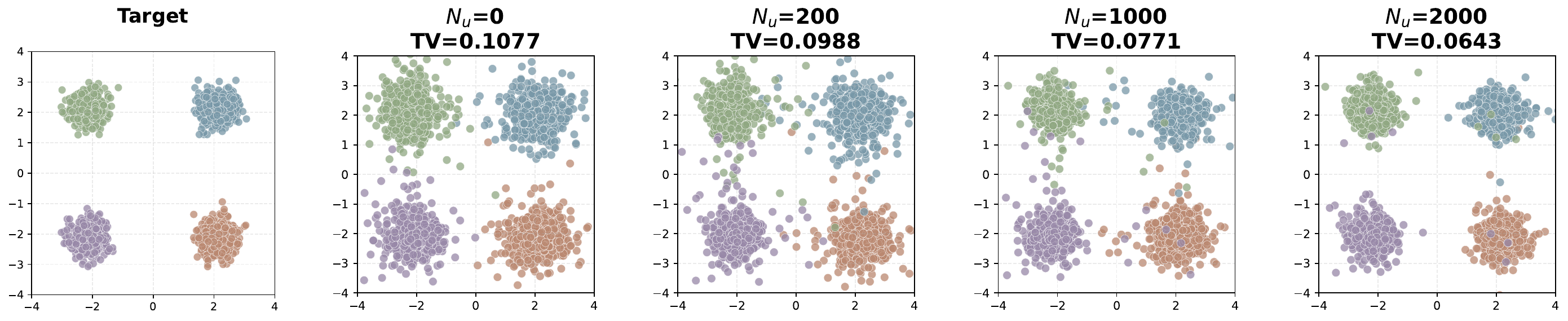}
	\includegraphics[width=0.95\textwidth]{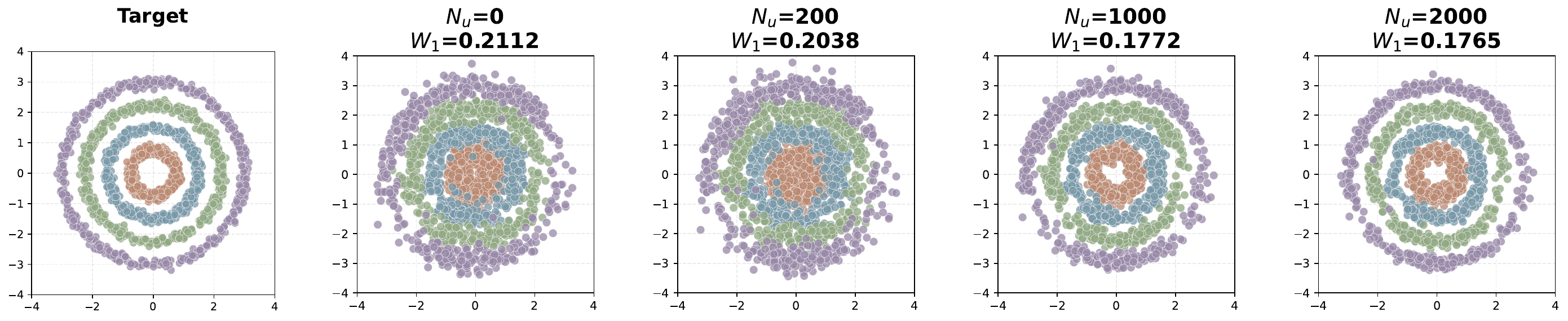}
	\includegraphics[width=0.95\textwidth]{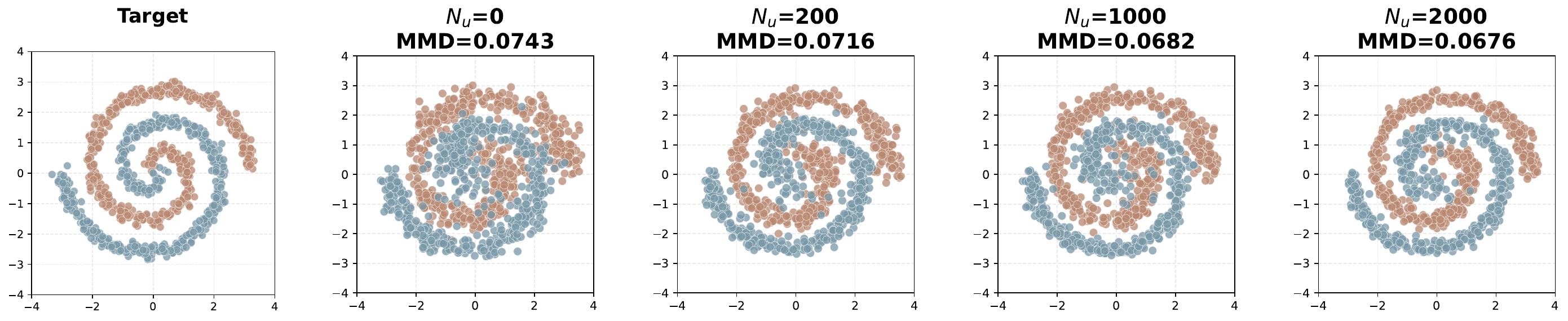}
	\caption{Target distributions and generated samples for \texttt{large\_4gaussians}, \texttt{rings}, and \texttt{2spirals} as $N_u$ increases. Different colors represent different classes. For each example,  $N_{\ell} = 200$  with varying $N_u \in \{0,200,\ldots,1{,}000,2{,}000\}$.
}
	\label{fig:selected_target_distributions_toy2d}
	\label{fig:toy2d_generated_samples_l4r2}
\end{figure}

\begin{figure}[htbp]
	\centering
	\includegraphics[width=0.3\textwidth]{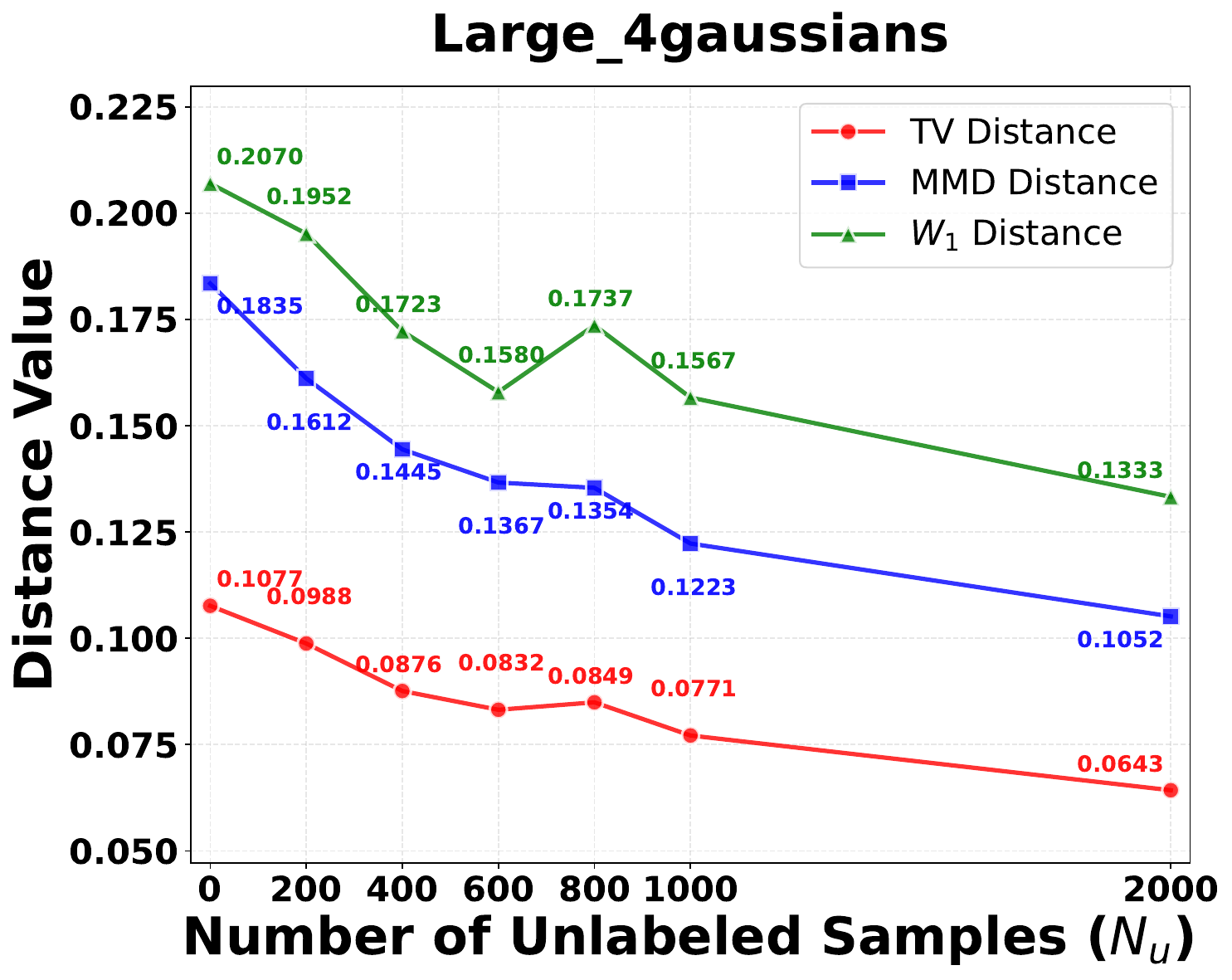}
	\includegraphics[width=0.3\textwidth]{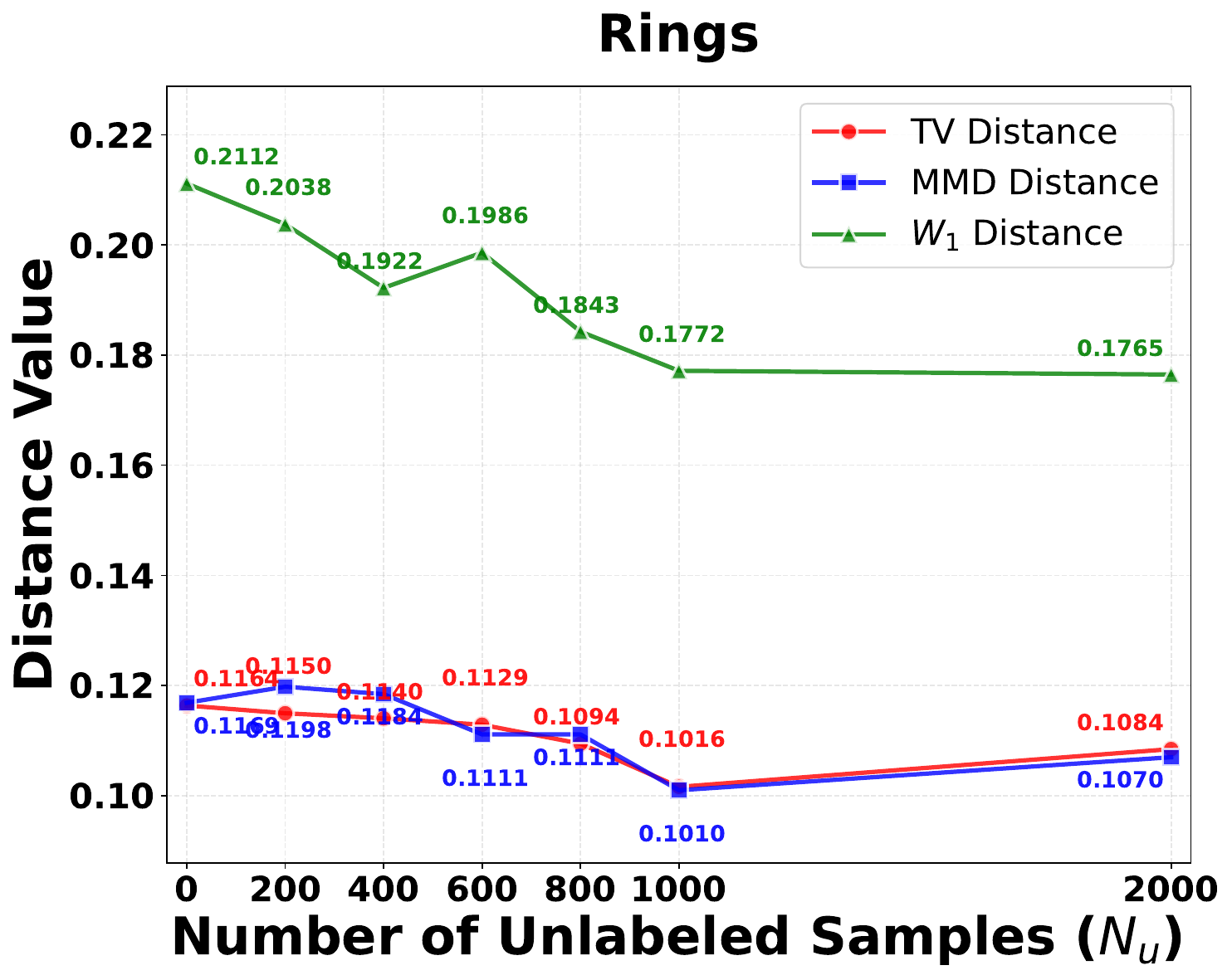}
	\includegraphics[width=0.3\textwidth]{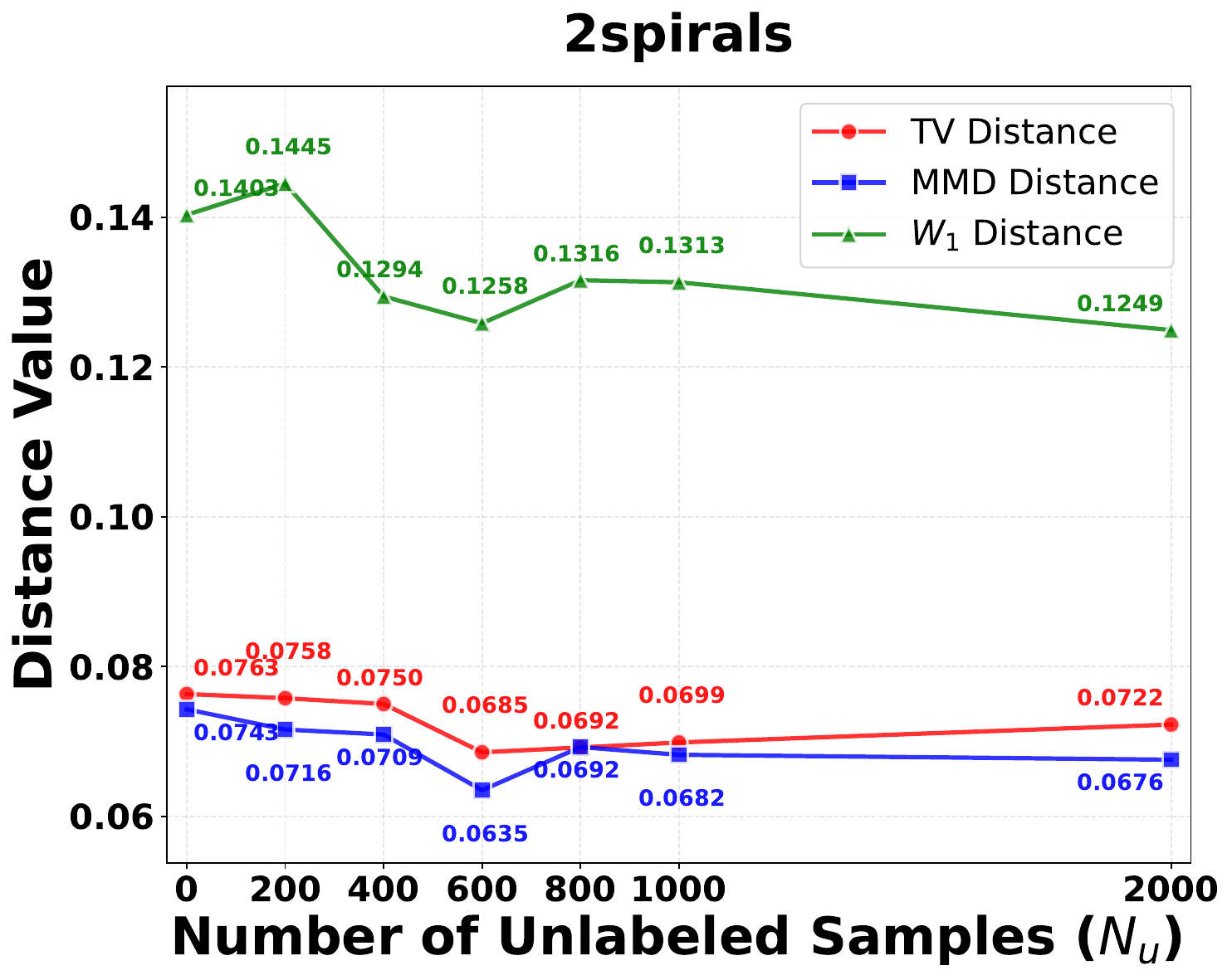}
	\caption{Class-averaged TV, MMD, and $W_1$ distances on \texttt{large\_4gaussians}, \texttt{rings}, and \texttt{2spirals} versus $N_u$. For each example,  $N_{\ell} = 200$  with varying $N_u \in \{0,200,\ldots,1{,}000,2{,}000\}$.
}
	\label{fig:toy2d_metric_curves}
\end{figure}

Figure~\ref{fig:toy2d_generated_samples_l4r2} displays the target distributions
together with generated samples across different values of $N_u$, while
Figure~\ref{fig:toy2d_metric_curves} reports the corresponding distance curves.
As $N_u$ increases, generated samples better recover the target
geometry: the Gaussian clusters become more compact and well separated, the
concentric rings become more clearly delineated, and the two spiral arms become
more distinct.
The corresponding TV, MMD, and $W_1$ curves generally decrease as $N_u$ increases, with mild fluctuations in some cases.
For \texttt{large\_4gaussians}, TV,
MMD, and $W_1$ decrease by approximately 40.3\%, 42.7\%, and 35.6\%,
respectively, as $N_u$ grows from $0$ to $2{,}000$.
The \texttt{rings} and \texttt{2spirals} examples also show overall decreasing trends, but the reductions are more modest than for \texttt{large\_4gaussians}.

These findings reflect the convergence bounds. The TV reductions align with the guaranteed strict improvement for discrete $Y$. For $W_1$, although unlabeled observations do not improve the asymptotic rate at $d_x = 2$, the finite-sample bound still contains terms decreasing with $N_u$, consistent with the observed downward trends.

\subsection{CIFAR-10 Dataset}
\label{sec:cifar10}

CIFAR-10\footnote{https://www.cs.toronto.edu/~kriz/cifar.html} \citep{krizhevsky2009learning} comprises $n_r = 60{,}000$ color
images ($x \in \mathbb{R}^{32 \times 32 \times 3}$) across $K = 10$ classes
with $n_{r,y} = 6{,}000$ images per class.
We consider $N_{\ell} \in \{1{,}000, 2{,}000, 5{,}000\}$ labeled samples,
with labels selected approximately uniformly across the ten classes, and
unlabeled-to-labeled ratios $\{0, 1, 2, 3, 4, 5, 10\}$.
The model architecture is a UNet with DDPM++ configuration under the EDM
framework \citep{karras2022elucidating}.
% with implementation details provided in the Supplementary Materials.

Generation quality is assessed via the Fr\'{e}chet Inception Distance
(FID) on $n_g = 60{,}000$ generated images
($n_{g,y} = 6{,}000$ per class).
FID evaluates image quality and diversity by comparing the feature distributions of real and generated images extracted from a pre-trained Inception network \citep{heusel2017gans}. We report two variants: overall FID,
$\text{FID} = \|\mu_r - \mu_g\|_2^2 +
\mathrm{Tr}\!\left(\Sigma_r + \Sigma_g - 2(\Sigma_r\Sigma_g)^{1/2}\right)$,
where $\mu_r, \mu_g$ and $\Sigma_r, \Sigma_g$ are the means and covariances of real and generated features; and class-averaged FID,
$\text{FID}_{\text{ca}} = \sum_{y \in \mathcal{Y}}\text{FID}(y)/K$,
where $\text{FID}(y)$ applies the above formula using class-specific
statistics $\mu_{r,y}, \mu_{g,y}, \Sigma_{r,y}, \Sigma_{g,y}$.
For each configuration, we report the best FID value attained during training.

\begin{figure}[t]
	\centering
	\includegraphics[width=0.35\textwidth]{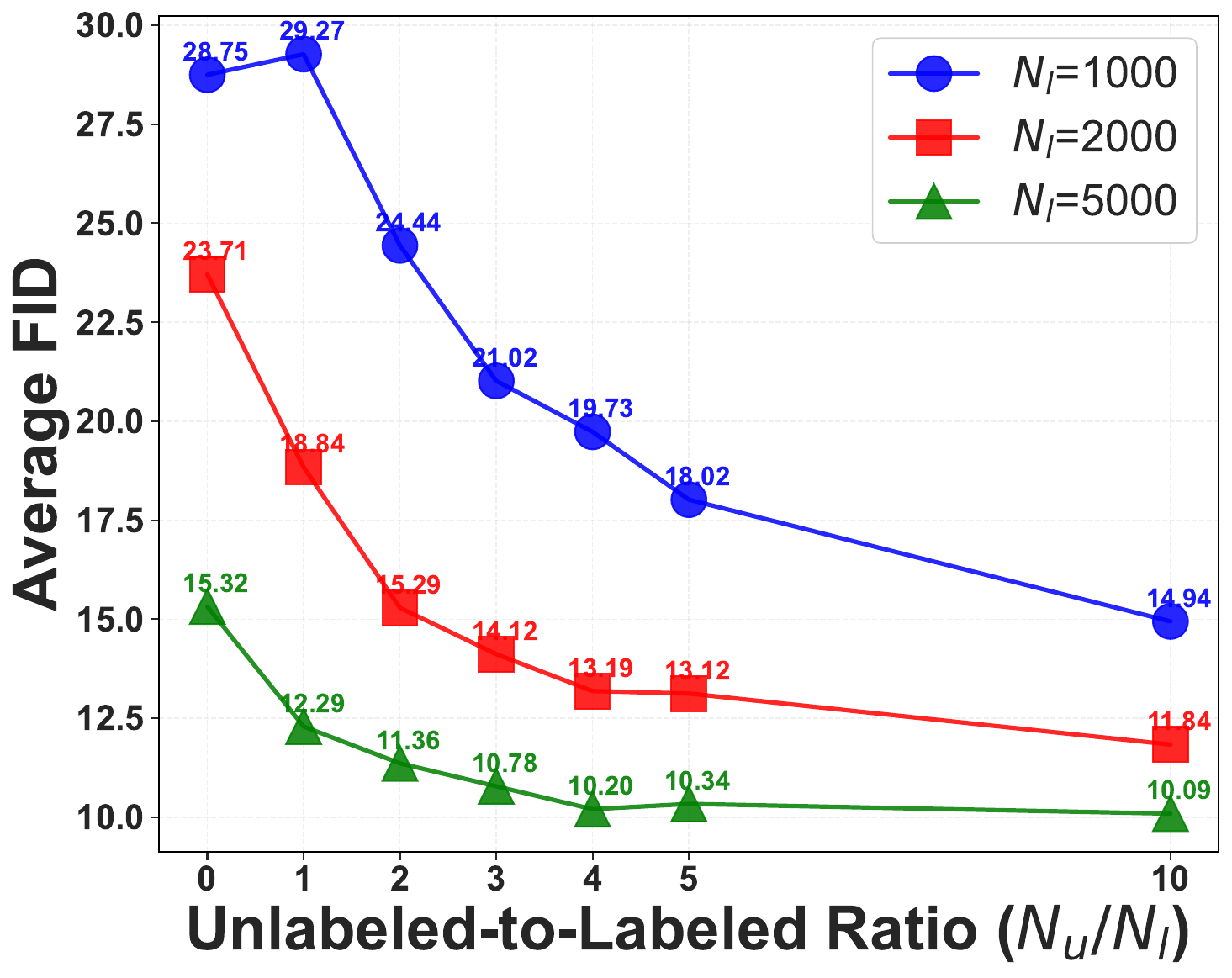}
	\includegraphics[width=0.35\textwidth]{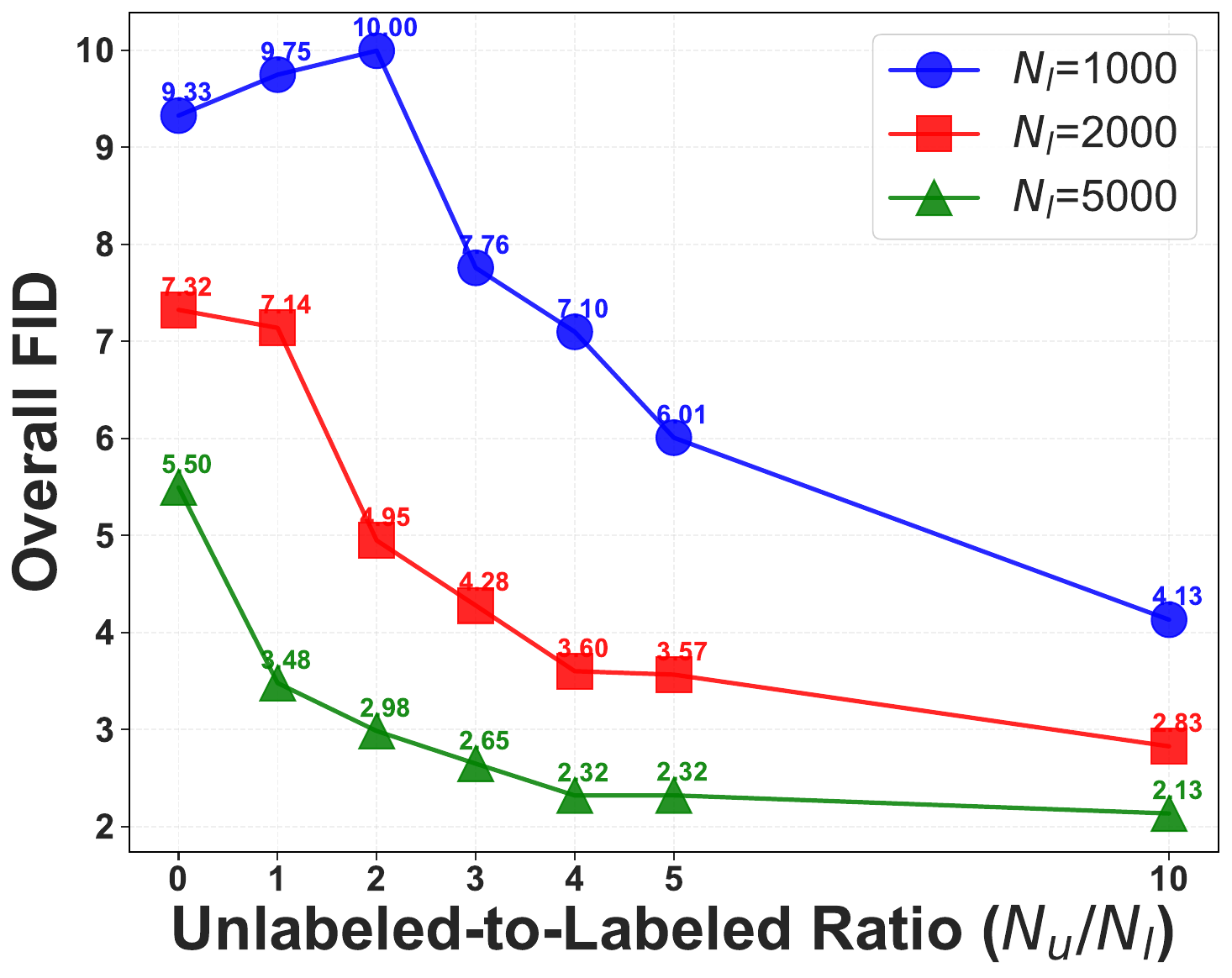}
		\caption{CIFAR-10: class-averaged FID (left) and overall FID (right) versus $N_u/N_{\ell}$ for $N_{\ell} \in \{1{,}000,2{,}000,5{,}000\}$.}
	\label{fig:fid_curves_cifar10}
\end{figure}
Figure~\ref{fig:fid_curves_cifar10} presents the best FID values achieved across different experimental configurations.
Both class-averaged and overall FID generally decrease as $N_u$ increases, with some fluctuations at small $N_u$ for $N_{\ell}=1{,}000$. The largest gains occur when unlabeled data are first added. Class-averaged FID decreases from 28.75 to 14.94 (48.0\%) for $N_{\ell} = 1{,}000$, from 23.71 to 11.84 (50.1\%) for $N_{\ell} = 2{,}000$, and from 15.32 to 10.09 (34.1\%) for $N_{\ell} = 5{,}000$. Overall FID decreases by 55.7\%, 61.3\%, and 61.3\%, respectively. Even at $N_{\ell} = 5{,}000$, additional unlabeled data further improve both FID measures.

Unlabeled data can also compensate for limited labeled samples:
$N_{\ell} = 1{,}000$ with $N_u = 10{,}000$ (Class-averaged FID: 14.94)
matches $N_{\ell} = 5{,}000$ alone (15.32), and $N_{\ell} = 2{,}000$ with
$N_u = 20{,}000$ (11.84) outperforms $N_{\ell} = 5{,}000$ with
$N_u = 5{,}000$ (12.29), consistent with the theoretical prediction that unlabeled data improve estimation through the shared parameter.
Notably, the fully supervised EDM baseline
uses all 50{,}000 training images (Overall FID: 1.68,
Class-averaged FID: 8.44), yet our method with only 10\% labels
($N_{\ell} = 5{,}000$) and $N_u = 50{,}000$ achieves Overall FID 2.13, Class-averaged FID: 10.09, closing most of this gap.

\begin{figure}[H]
	\centering
	\includegraphics[width=0.4\textwidth]{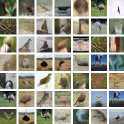}
\hspace{0.05\textwidth}
	\includegraphics[width=0.4\textwidth]{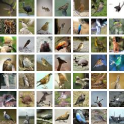}
	\caption{CIFAR-10 generated samples conditioned on the class ``bird'' under
		$N_{\ell}=1{,}000$. Left: $N_u=0$; right: $N_u=10{,}000$. }
	\label{fig:cifar10_generated_samples}
\end{figure}

Since some limited-label configurations achieve competitive FID scores, we
further examine whether these scores are accompanied by visually meaningful generation rather than potential memorization of the labeled training samples.
Figure~\ref{fig:cifar10_generated_samples} shows generated samples conditioned on the class ``bird'' under $N_{\ell}=1{,}000$, comparing $N_u=0$ with $N_u=10{,}000$. With labeled data alone, the generated samples contain more repeated shapes and less varied backgrounds (e.g., the third sample in the fourth row and the second sample in the seventh row have very similar elongated bird silhouettes), whereas adding unlabeled data produces more diverse poses, colors, and
backgrounds.
As additional diagnostics, we also examine t-SNE projections of Inception V3 features and FID training curves.
% both reported in the Supplementary Materials.
The t-SNE visualization compares real data, fully supervised (FS), and
semi-supervised (SS) outputs for the class ``horse'', showing that the SS
embeddings better cover the regions occupied by real-data embeddings across all
$N_{\ell}$ settings. The training curves show that, in settings with
adequate data, larger datasets tend to reach their minimum FID values later
(measured in kimg, i.e., thousands of processed images), reflecting the larger
number of training images processed. In contrast, several low-label
configurations show less regular training behavior: the FID curves either keep
decreasing over the recorded training window or attain a minimum and then rise,
including $N_{\ell} = 1{,}000$ with $N_u \in \{0, 1{,}000, 2{,}000\}$ and
$N_{\ell} = 2{,}000$ with $N_u = 0$.
Together, these diagnostics suggest that unlabeled data improve feature-space coverage and reduce the memorization in low-label regimes.

\subsection{Intel Scenes Dataset}
\label{sec:intel6}
The Intel Scenes dataset\footnote{https://www.kaggle.com/datasets/puneet6060/intel-image-classification}\citep{puneet2019intel} comprises $n_r = 17{,}034$
natural landscape images ($x \in \mathbb{R}^{64 \times 64 \times 3}$) across
$K = 6$ categories: buildings (2,628), forest (2,745), glacier (2,957),
mountain (3,037), sea (2,784), and street (2,883). With higher resolution
and more diverse scene categories, ranging from natural landscapes to urban
environments, Intel Scenes presents a distinct challenge from CIFAR-10.
% and AFHQv2.
We consider $N_{\ell} \in \{1{,}200, 2{,}400\}$ with unlabeled-to-labeled
ratios $\{0, 1, 2, 3, 4, 5, 10\}$ for $N_{\ell} = 1{,}200$ and $\{0, 1, 2, 3, 4, 5\}$
for $N_{\ell} = 2{,}400$, generating $n_{g,y} = 3{,}000$ images per category
and evaluating the generated samples using FID as defined in Section~\ref{sec:cifar10}.

\begin{figure}[t]
	\centering
	\includegraphics[width=0.35\textwidth]{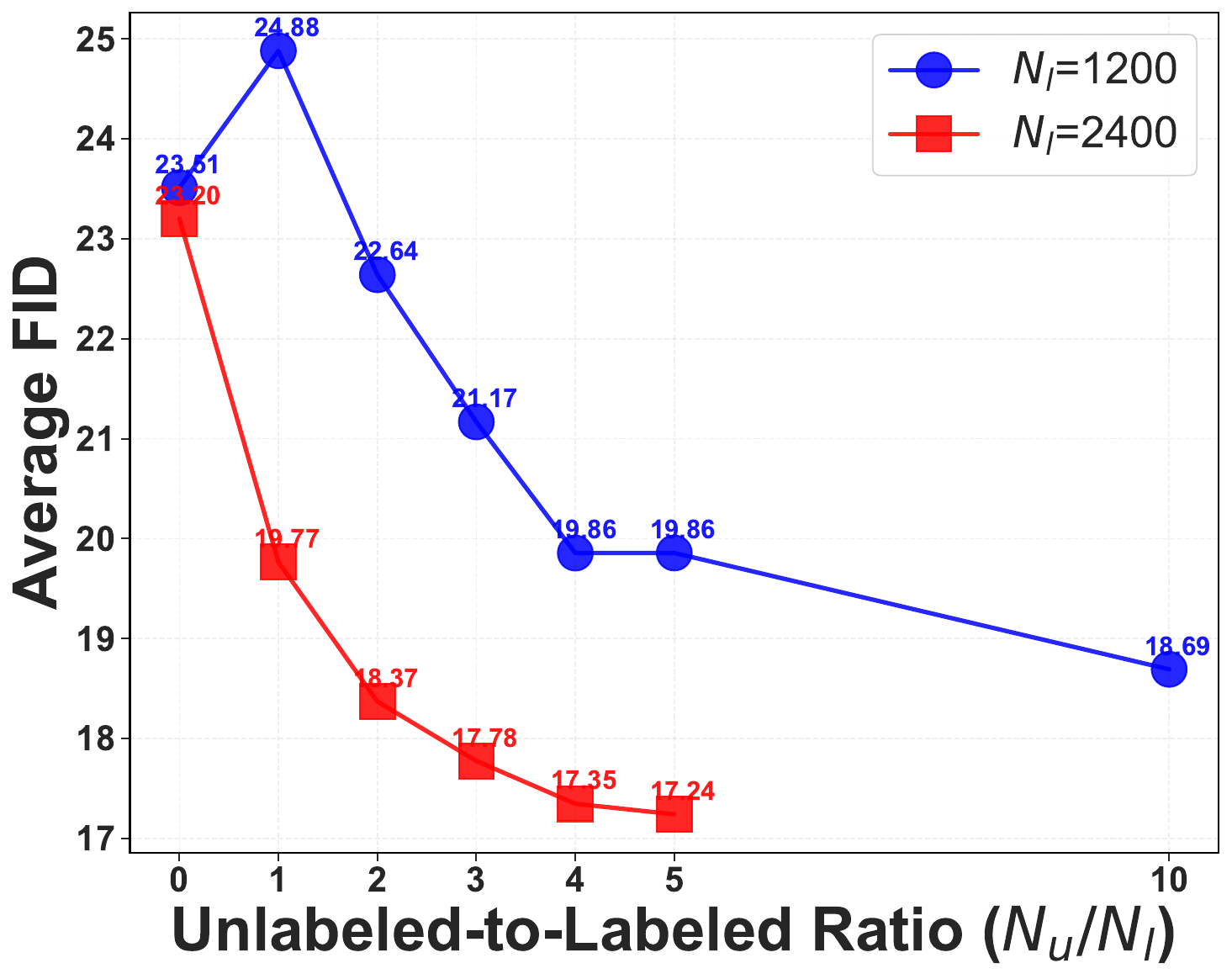}
	%\hspace{0.03\textwidth}
	\includegraphics[width=0.35\textwidth]{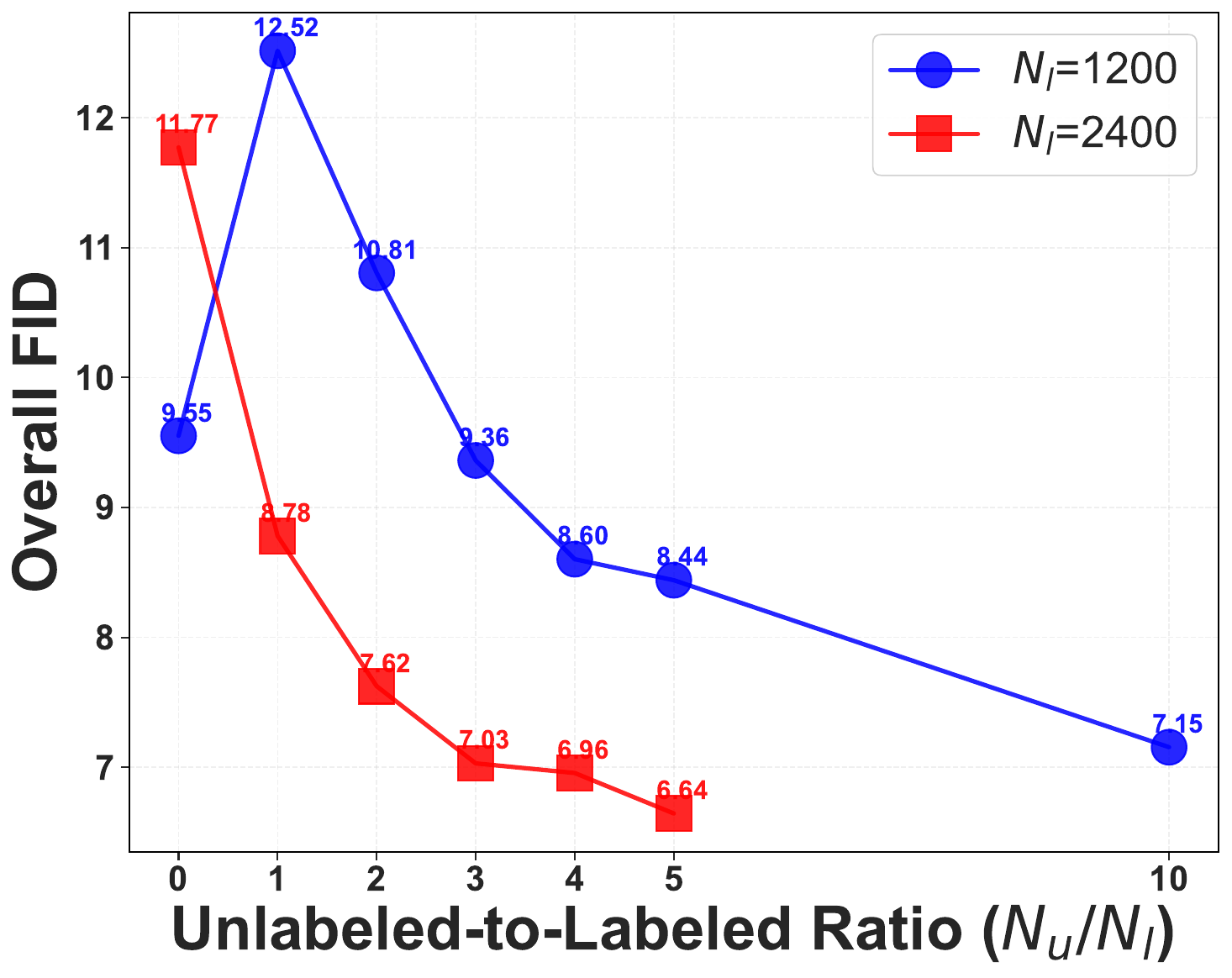}
	\caption{Intel Scenes: class-averaged FID (left) and overall FID (right) versus $N_u/N_{\ell}$ for $N_{\ell} \in \{1{,}200,2{,}400\}$.}
	\label{fig:fid_curves_intel6}
\end{figure}

\begin{figure}[ht]
	\centering
	\includegraphics[width=0.2\textwidth]{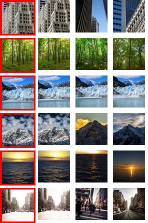}
	\hspace{0.5em}
	\includegraphics[width=0.2\textwidth]{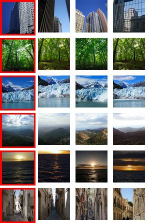}
	\hspace{1.5em}
	\includegraphics[width=0.2\textwidth]{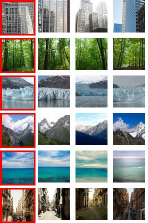}
	\hspace{0.5em}
	\includegraphics[width=0.2\textwidth]{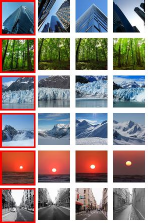}
	\caption{Intel Scenes: CLIP-based nearest-neighbor analysis for generated images. Generated images (red borders) and their closest training-set matches are shown. Left to right: ($N_{\ell}=1{,}200, N_u=0$), ($N_{\ell}=1{,}200, N_u=6{,}000$), ($N_{\ell}=2{,}400, N_u=0$), ($N_{\ell}=2{,}400, N_u=12{,}000$).}
	\label{fig:intel6_clip_nearest_neighbor}
\end{figure}

Figure~\ref{fig:fid_curves_intel6} presents FID results on Intel Scenes.
Both metrics generally decrease as the unlabeled-to-labeled ratio increases.
Specifically, the class-averaged
FID decreases from 23.51 to 18.69 (20.5\%) for $N_{\ell} = 1{,}200$
and from 20.31 to 15.12 (25.6\%) for $N_{\ell} = 2{,}400$. The overall FID shows
similar decreases, from 9.55 to 7.15 (25.1\%) for
$N_{\ell} = 1{,}200$ and from 7.62 to 5.49 (28.0\%) for $N_{\ell} = 2{,}400$.
At $N_{\ell} = 1{,}200$, both metrics fluctuate at small $N_u$ before declining more steadily once $N_u \geq 2{,}400$.
We further examine sample similarity using a CLIP-based nearest-neighbor analysis. Generated and training images are embedded with CLIP, and each generated image is paired with its nearest training-set match by feature inner product. Figure~\ref{fig:intel6_clip_nearest_neighbor} complements the FID analysis. At $N_{\ell} = 1{,}200$ with $N_u = 0$, generated images closely resemble training samples, whereas the corresponding samples with unlabeled data show greater visual variation. At $N_{\ell} = 2{,}400$ with $N_u = 0$, generated images already differ more noticeably from their nearest training neighbors, consistent with the more stable FID decreases for $N_{\ell} = 2{,}400$.

\subsection{Bangalore EEG Epilepsy Dataset}
\label{sec:beed}
The Bangalore EEG Epilepsy Dataset (BEED)\footnote{https://www.kaggle.com/datasets/mdnaim/eeg-epilepsy-beed } %\footnote{https://archive.ics.uci.edu/ml/datasets/BEED\%3A+Bangalore+EEG+Epilepsy+Dataset}
\citep{beed2025} is a
tabular dataset comprising $8{,}000$ samples with 16-dimensional EEG
feature vectors, evenly distributed across four epilepsy-related
classes (2,000 per class): healthy (0), generalized seizures (1),
focal seizures (2), and seizure events (3).
We partition the data into training (75\%), validation (10\%), and test (15\%) sets.
From the training set, we use $N_{\ell}=800$ labeled samples and vary
$N_u \in \{0,800,1600,2400,3200,4000,5000\}$ unlabeled samples.
We also
include a fully supervised reference with $N_{\ell}=6{,}000$, corresponding to all training samples being labeled. The best model is selected by sliced $W_1$ distance on the validation set.
For final evaluation, we generate samples matching the test-set class sizes.
%Network architecture and training details are provided in the Supplementary Materials.

We use the following four evaluation metrics.
\begin{itemize}
\item[(a)] Feature-wise Wasserstein-1 distance: averages the
one-dimensional Wasserstein distances over the 16 features.

\item[(b)] Sliced Wasserstein-1 distance: measures overall
distributional similarity through random one-dimensional projections;
see Section~\ref{sec:simulation} for the definition. % We use $K = 1{,}000$ random projections.

\item[(c)] KS Complement: Measures marginal distribution
alignment for each feature dimension,
\begin{align*}
\text{KSComp} = \frac{1}{d}\sum_{j=1}^d
\left(1 - \sup_{t}|F_{r,j}(t) - F_{g,j}(t)|\right) \in [0, 1],
\end{align*}
where $d = 16$ is the feature dimension and $F_{r,j}$, $F_{g,j}$ are
empirical CDFs of real and generated data for feature $j$.
\item[(d)] Boundary Adherence (BA): Measures whether generated
samples fall within the value ranges of real data,
\begin{align*}
\text{BA} = \frac{1}{d}\sum_{j=1}^d \frac{1}{n_g}\sum_{i=1}^{n_g}
\mathbf{1}\left\{x_{g,i,j} \in [\min_k x_{r,k,j},
\max_k x_{r,k,j}]\right\} \in [0, 1],
\end{align*}
where $\{x_{r,k}\}_{k=1}^{n_r}$ are real samples and $\{x_{g,i}\}_{i=1}^{n_g}$ are generated samples.
\end{itemize}
Each metric is computed class-conditionally and then averaged across the four classes.
Results are reported as mean $\pm$ standard deviation over five random seeds.

\begin{figure}[t]
\centering
\includegraphics[width=0.95\textwidth]{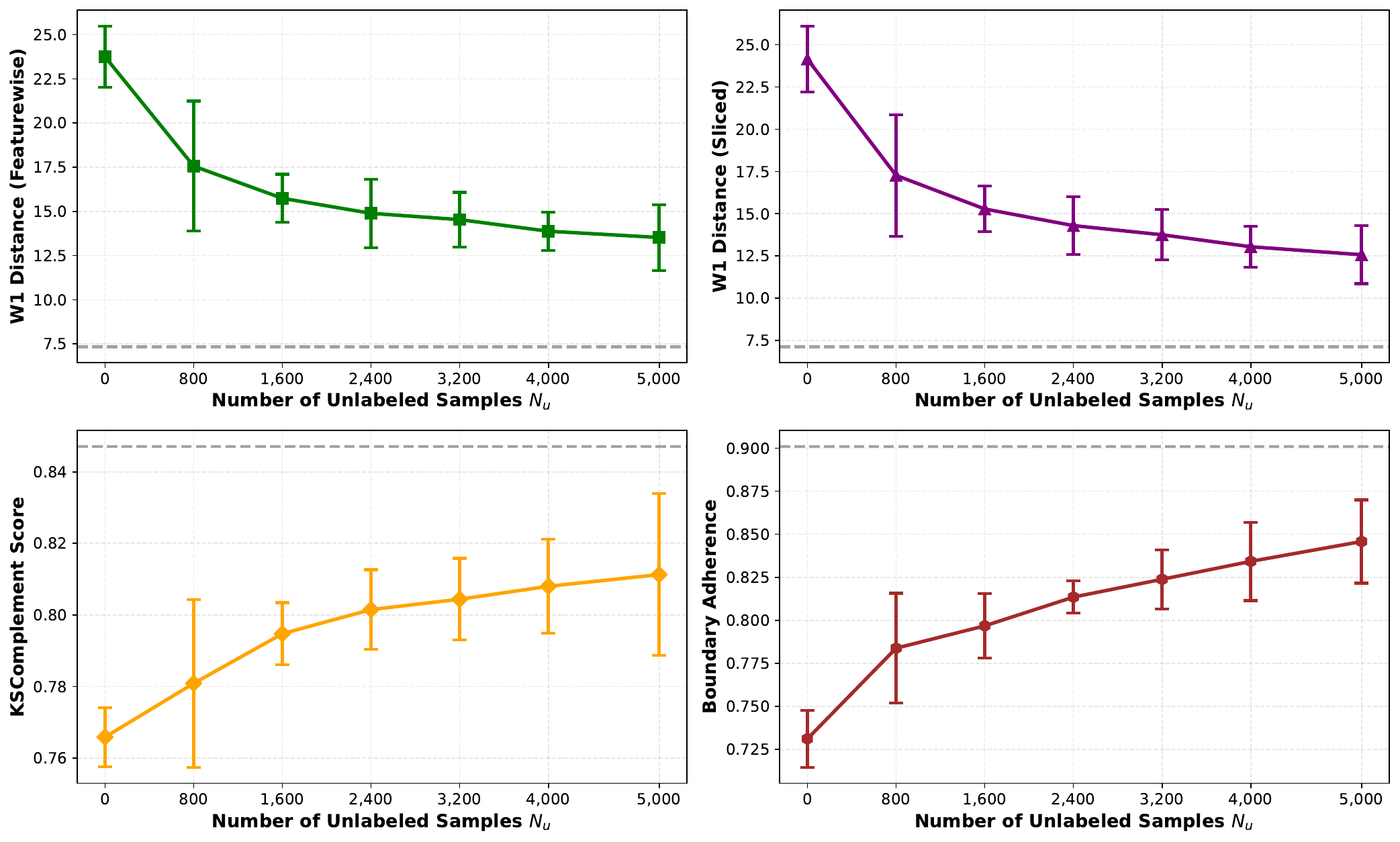}
\caption{BEED results with $N_{\ell}=800$. Curves and error bars show mean $\pm$ standard deviation over five random seeds. Grey dashed lines indicate the reference obtained by using all training samples as labeled data. Arrows in the axis labels indicate the favorable direction.}
\label{fig:beed_metrics}
\end{figure}

Figure~\ref{fig:beed_metrics} shows an overall improvement as $N_u$ increases.
The two $W_1$ metrics decrease, indicating that the generated distribution becomes closer to the real data distribution. At the same time, increases in KSComp and BA indicate better agreement in the marginal feature distributions and a reduced tendency to generate implausible feature values, respectively. More specifically, the amount of improvement is not uniform across the range of $N_u$, with larger gains at smaller values of $N_u$ and more gradual changes thereafter.
This suggests diminishing marginal gains from additional unlabeled samples.
Relative to the fully supervised reference, the results with $N_{\ell}=800$ move closer to the reference levels as $N_u$ increases, although a gap remains. In this sense, adding unlabeled covariates partly offsets the degradation caused by having far fewer labeled samples. Together, these results provide a tabular-data example in which unlabeled covariates improve conditional generation under label scarcity.

\section{Conclusion}
\label{sec:Conclusion}
In this paper, we propose LACD, a semi-supervised diffusion model for conditional generation via label augmentation. We establish finite-sample convergence rates in $W_1$ and TV distances,
demonstrating theoretical improvements over the supervised counterpart.
Numerical experiments on synthetic, image, and tabular benchmarks support these theoretical properties and show that unlabeled data can improve finite-sample performance in conditional generative modeling. Several promising directions deserve further investigation. On the theoretical side, establishing minimax lower bounds to complement the upper bounds would confirm whether the proposed rates are optimal. Furthermore, under scenarios where $X$ or $Y$ admits a low-dimensional structure such as a manifold, deriving sharper rates that replace the ambient dimension with the intrinsic dimension offers a compelling theoretical extension. Beyond these theoretical questions, relaxing the data assumptions to allow label missingness to depend on $X$ or $Y$ would broaden the practical scope of the framework. On the methodological front, adapting the shared-parameter design
to other generative frameworks, such as normalizing flows or GANs, could broaden the approach's applicability. Finally, using large language models to generate pseudo-labels for unlabeled data is a promising direction for future work.

\section*{Competing interests}
No competing interest is declared.

\begin{singlespace}
\bibliographystyle{apalike} %abbrvnat}
\bibliography{LACDarXiv0718}
\end{singlespace}

\end{document}